\documentclass[runningheads]{llncs}

\usepackage{eccv}

\usepackage{eccvabbrv}

\usepackage{graphicx}

\usepackage[accsupp]{axessibility}  

\usepackage{hyperref}

\usepackage[dvipsnames]{xcolor}

\usepackage[capitalize]{cleveref}
\makeatletter
\AddToHook{cmd/appendix/before}{\def\cref@section@alias{appendix}}
\makeatother
\crefname{section}{Sec.}{Secs.}
\Crefname{section}{Section}{Sections}
\Crefname{table}{Table}{Tables}
\crefname{table}{Tab.}{Tabs.}
\Crefname{append}{Appendix}{Appendixs}
\crefname{append}{Append.}{Appends.}
\Crefname{subfigure}{Figure}{Figures}
\crefname{subfigure}{Fig.}{Figs.}

\usepackage{mathtools}
\usepackage{amsmath}

\usepackage{booktabs}       
\usepackage{arydshln}
\usepackage{multirow}
\usepackage{tabularx}
\usepackage{array}
\usepackage{colortbl}       
\usepackage{afterpage}

\usepackage{graphicx}
\usepackage{subcaption}
\usepackage{rotating}
\usepackage{float}
\usepackage{tikz}
\usepackage{pgffor}
\usepackage{placeins}

\usepackage{algorithm}
\usepackage{algpseudocode}

\usepackage[accsupp]{axessibility}

\newcommand{\redbox}[1]{%
  \begin{tikzpicture}[baseline=(img.south)]
    \node[inner sep=0pt] (img) {#1};
    \draw[red, line width=2pt] (img.north west) rectangle (img.south east);
  \end{tikzpicture}%
}
\newcolumntype{C}{>{\centering\arraybackslash}X}
\definecolor{lightgray}{gray}{0.9}
\newcommand{\tightbold}[1]{{\fontseries{b}\selectfont #1}}

\usepackage{pifont}         
\usepackage{adjustbox}
\definecolor{darkgreen}{rgb}{0.0, 0.5, 0.0}

\hypersetup{
    colorlinks=true,
    linkcolor=magenta,
}

\usepackage{orcidlink}

\begin{document}

\title{VIGOR: VIdeo Geometry-Oriented Reward for Temporal Generative Alignment}

\titlerunning{VIdeo Geometry-Oriented Reward for Temporal Generative Alignment}

\author{Tengjiao Yin\inst{1}\orcidlink{0009-0004-7320-591X} \and
Jinglei Shi\inst{1}$^\dagger$\orcidlink{0000-0003-2926-0415} \and
Heng Guo\inst{2}\orcidlink{0000-0003-0047-3927} \and
Xi Wang\inst{3}\orcidlink{0000-0001-6586-1926}}

\authorrunning{T. Yin et al.}

\institute{VCIP \& TMCC \& DISSec, College of Computer Science, Nankai University \and
Beijing University of Posts and Telecommunications \and
LIX, \'Ecole Polytechnique, IP Paris}

\maketitle

\renewcommand{\thefootnote}{$^\dagger$}
\footnotetext{Corresponding author.}
\renewcommand{\thefootnote}{\arabic{footnote}}

\begin{center}
  \textcolor{magenta}{\url{https://vigor-geometry-reward.com/}}
\end{center}

\begin{abstract}
Video diffusion models lack explicit geometric supervision during training, leading to inconsistency artifacts such as object deformation, spatial drift, and depth violations in generated videos. To address this limitation, we propose a geometry-based reward model that leverages pretrained geometric foundation models to evaluate multi-view consistency through cross-frame reprojection error. Unlike previous geometric metrics that measure inconsistency in pixel space, where pixel intensity may introduce additional noise, our approach conducts error computation in a pointwise fashion, yielding a more physically grounded and robust error metric. Furthermore, we introduce a geometry-aware sampling strategy that filters out low-texture and non-semantic regions, focusing evaluation on geometrically meaningful areas with reliable correspondences to improve robustness. We apply this reward model to align video diffusion models through two complementary pathways: post-training of a bidirectional model via SFT or Reinforcement Learning and inference-time optimization of a Causal Video Model (e.g., Streaming video generator) via test-time scaling with our reward as a path verifier. Experimental results validate the effectiveness of our design, demonstrating that our geometry-based reward provides superior robustness compared to other variants. By enabling efficient inference-time scaling, our method offers a practical solution for enhancing open-source video models without requiring extensive computational resources for retraining. 

\end{abstract}
\section{Introduction}

\begin{figure}[t]
\centering
\def\imgwd{0.25\linewidth}
\begin{minipage}[c]{\textwidth}
    \includegraphics[width=\imgwd]{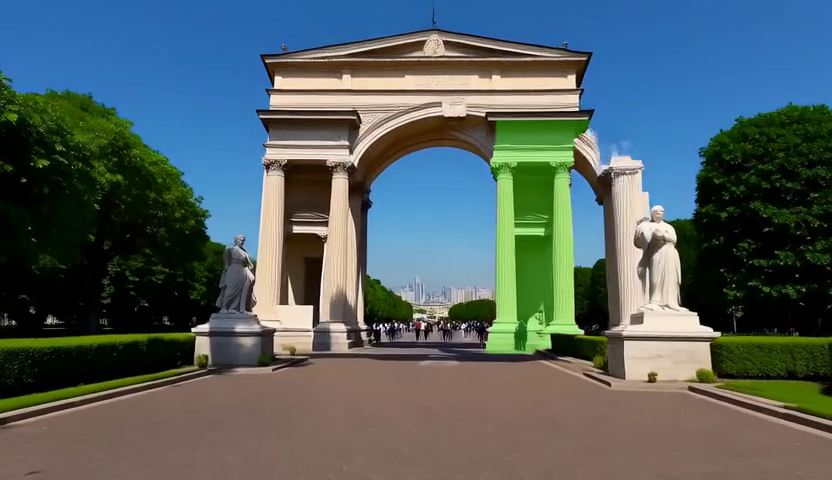}%
    \includegraphics[width=\imgwd]{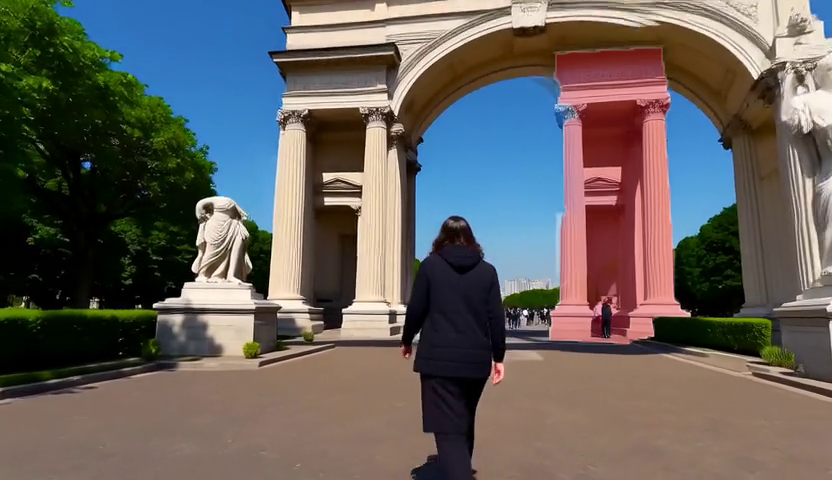}%
    \includegraphics[width=\imgwd]{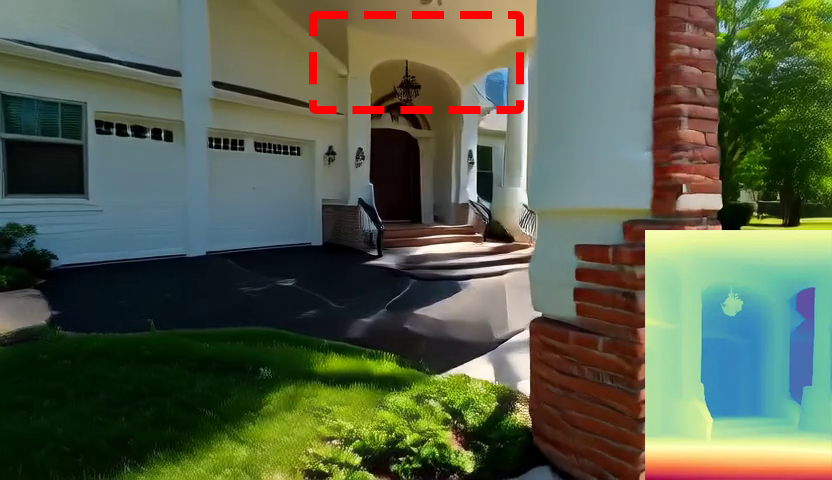}%
    \includegraphics[width=\imgwd]{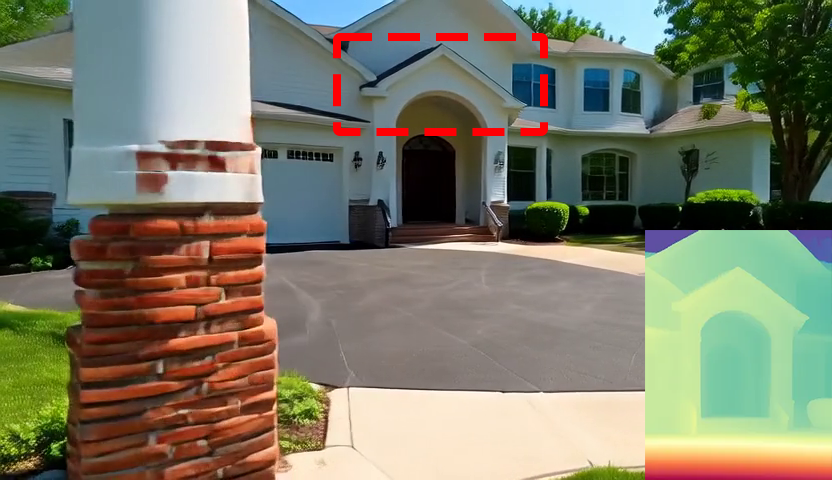}%
\end{minipage}

\begin{minipage}[c]{\textwidth}
    \centering\footnotesize Object Deformation\qquad\qquad\qquad\qquad\qquad Depth Violation
\end{minipage}

\begin{minipage}[c]{\textwidth}
    \includegraphics[width=\imgwd]{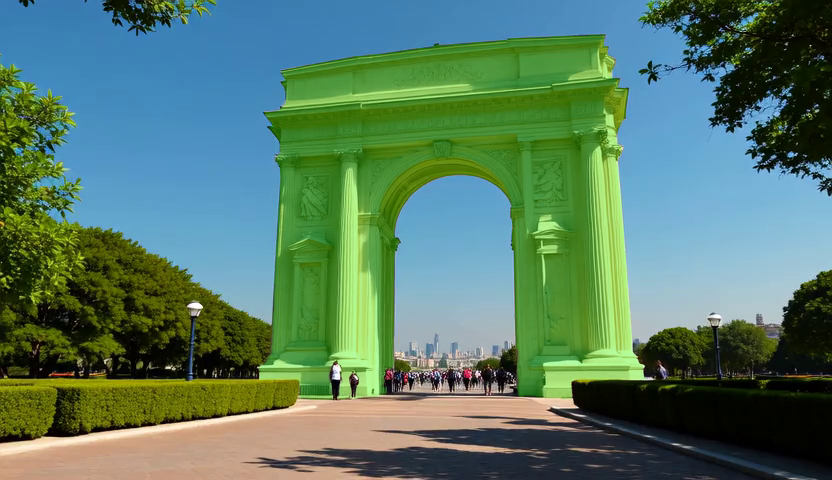}%
    \includegraphics[width=\imgwd]{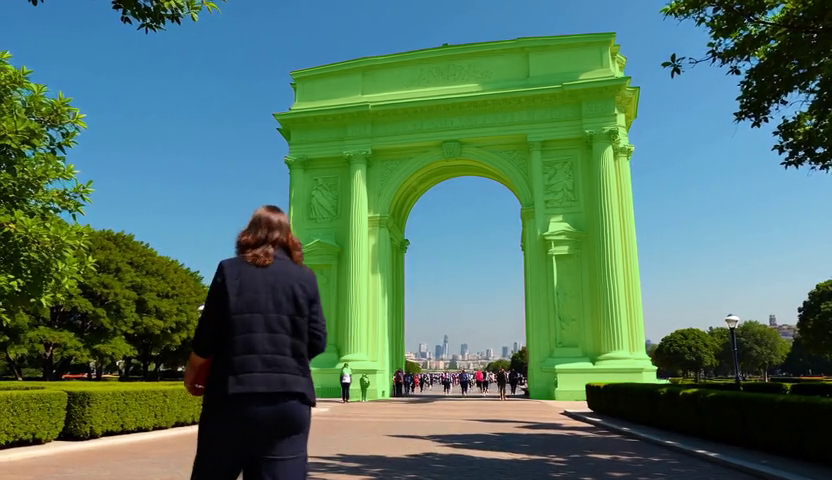}%
    \includegraphics[width=\imgwd]{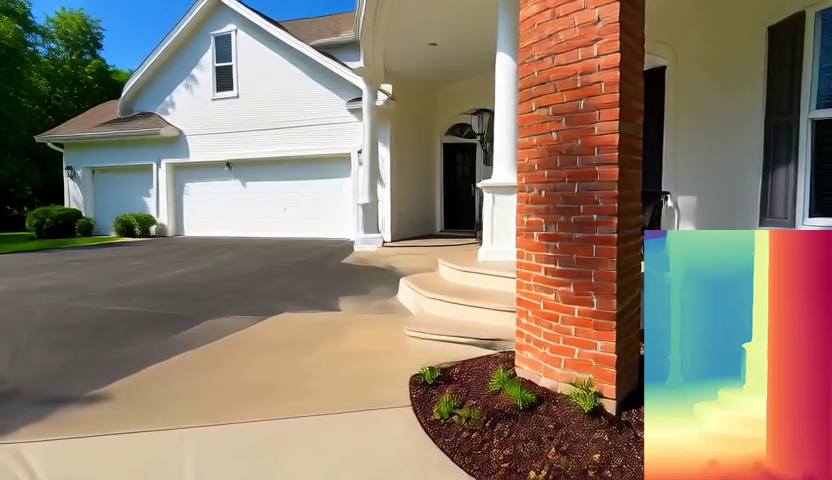}%
    \includegraphics[width=\imgwd]{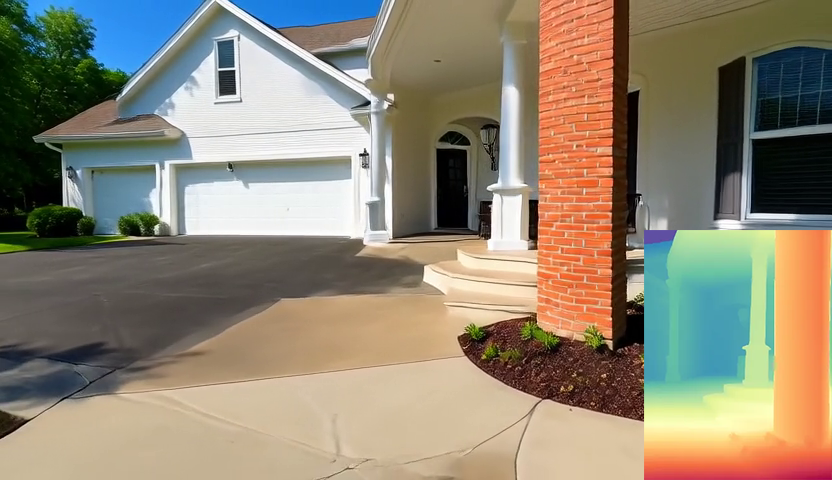}%
\end{minipage}
\begin{minipage}[c]{\textwidth}
    \centering\footnotesize  Improved Geometry (\textbf{Ours})
\end{minipage}

\begin{minipage}[c]{\textwidth}
    \includegraphics[width=\imgwd]{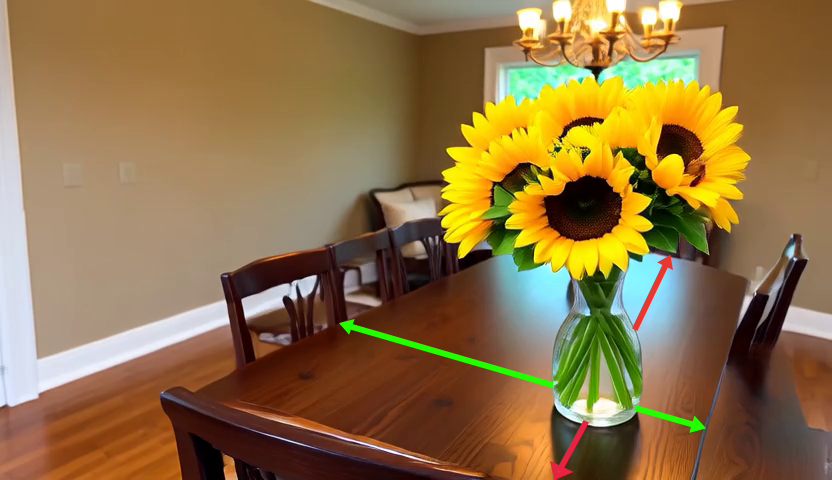}%
    \includegraphics[width=\imgwd]{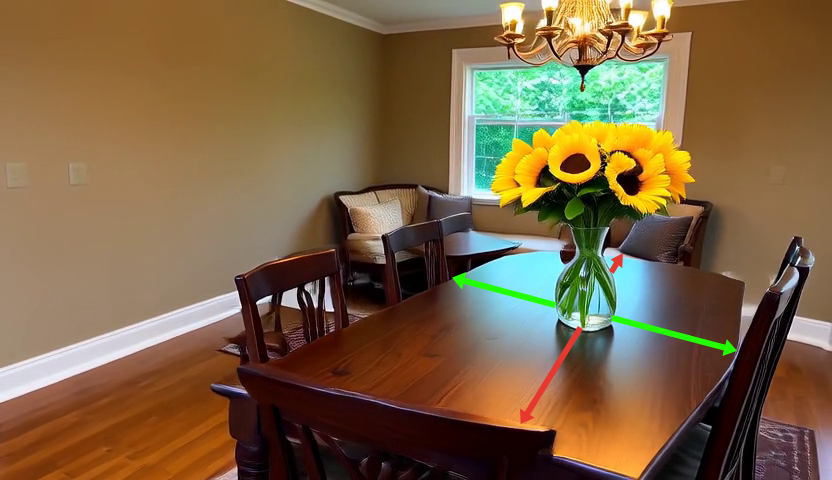}%
    \includegraphics[width=\imgwd]{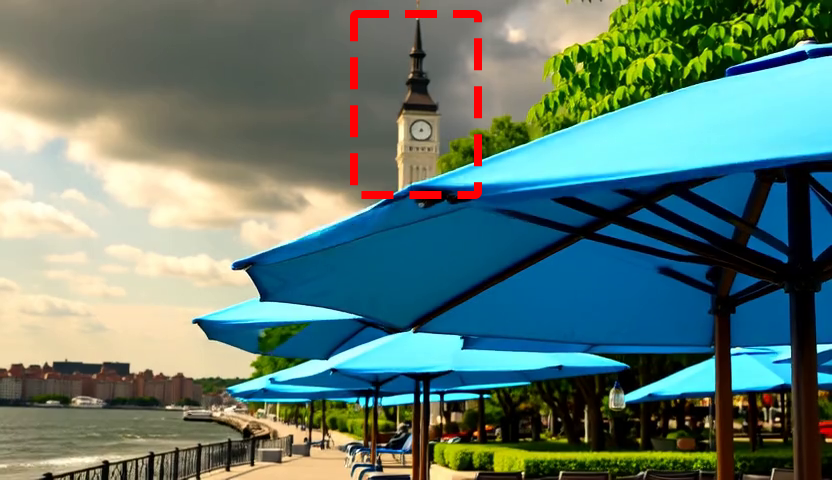}%
    \includegraphics[width=\imgwd]{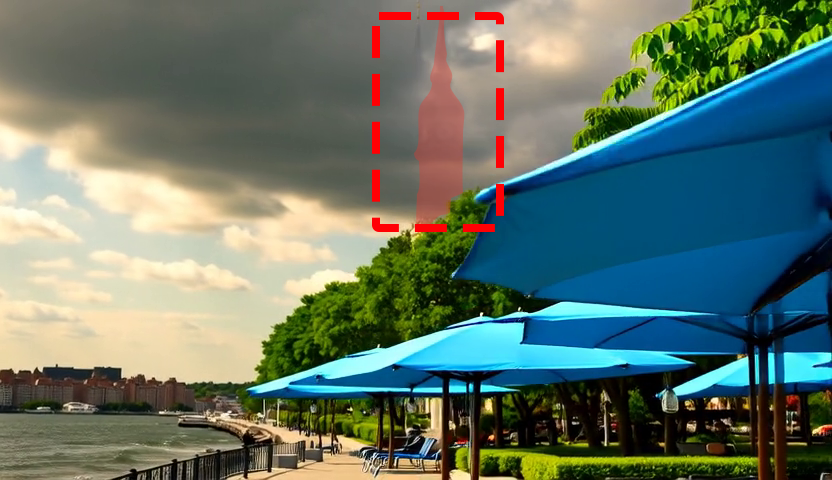}%
\end{minipage}
\begin{minipage}[c]{\textwidth}
    \centering\footnotesize Spatial Drift \qquad\qquad\qquad\qquad\qquad\qquad\qquad Flickering \qquad
\end{minipage}

\begin{minipage}[c]{\textwidth}
    \includegraphics[width=\imgwd]{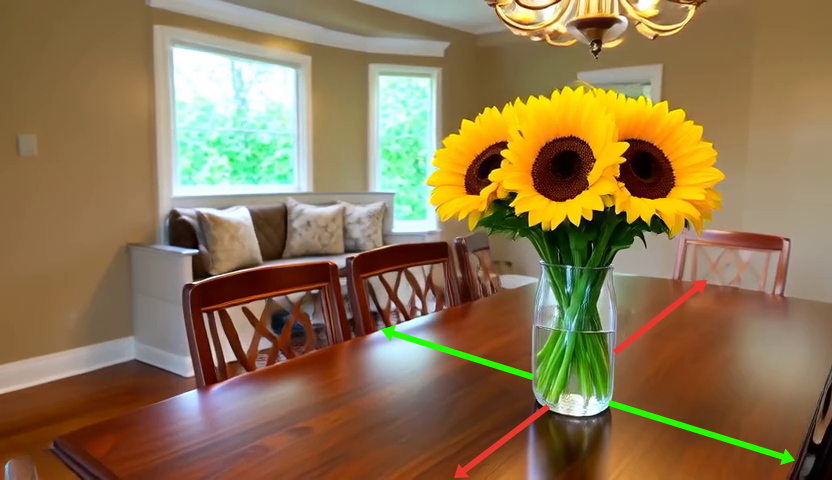}%
    \includegraphics[width=\imgwd]{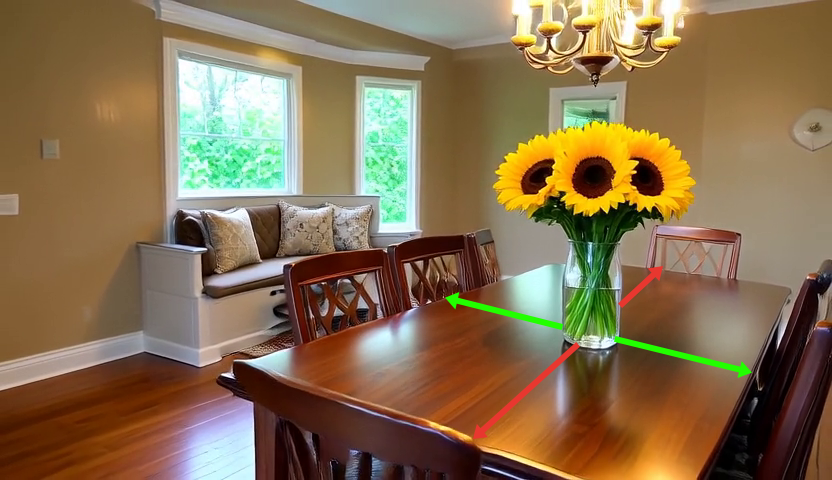}%
    \includegraphics[width=\imgwd]{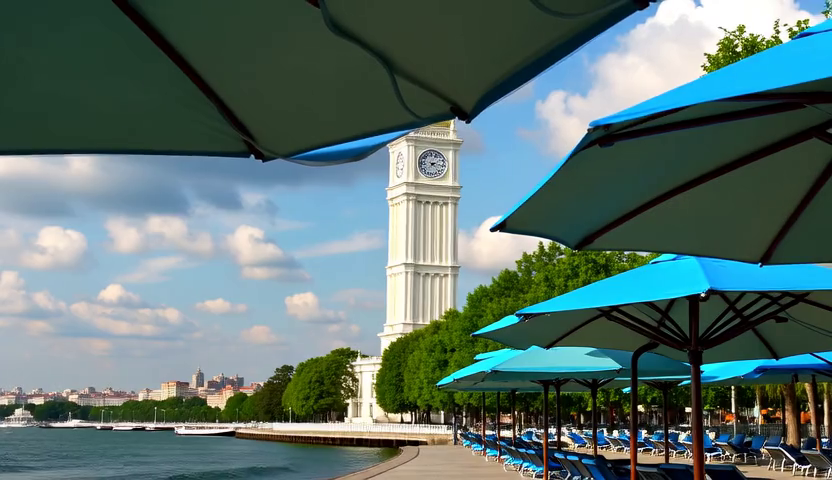}%
    \includegraphics[width=\imgwd]{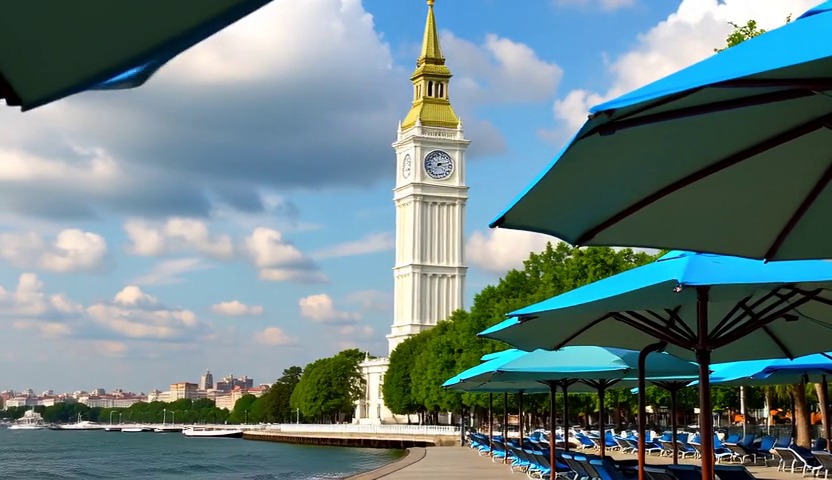}%
\end{minipage}
\begin{minipage}[c]{\textwidth}
    \centering\footnotesize  Improved Geometry (\textbf{Ours})
\end{minipage}

\caption{\textbf{Examples} of geometry artifacts in generated videos and our results.}
\label{fig:teaser}
\end{figure}

Video diffusion models have achieved remarkable progress in recent years~\cite{brooks2024sora,googledeepmind2025veo3,polyak2025moviegen,wan2025wan,kong2025hunyuanvideo,yang2025cogvideox} demonstrating increasingly photorealistic and temporally coherent video synthesis. Yet a fundamental limitation persists across all of these systems: \textit{a lack of explicit geometric supervision during training}. This results in emerging geometric artifacts (as shown in Fig.~\ref{fig:teaser}): object deformation, spatial drift, flickering, depth violations, physically implausible perspective changes, etc.

While closed-source models~\cite{brooks2024sora,googledeepmind2025veo3,polyak2025moviegen} tend to exhibit stronger geometric realism than their open-source counterparts, this advantage is largely attributable to training at an unprecedented scale: by fitting massive corpora of natural video, these models allow implicit geometric priors to emerge as a byproduct of sheer data volume. Some recent works have attempted a more principled alternative by incorporating explicit geometric supervision such as conditioning on depth maps~\cite{xing2023makeyourvideo,huang2025voyager,dai2025fantasyworld} or camera poses~\cite{wang2024motionctrl,he2025cameractrlii,wang2024akira}. However, such approaches are fundamentally constrained by data accessibility: geometric ground truth of this kind is rarely paired with the Internet-scale corpora on which modern video generators are trained, rendering explicit supervision impractical for open-source research and infeasible for customized downstream applications.

In light of these limitations, an increasingly adopted strategy for improving generative diffusion models is to align or tilt them toward an informative reward (either a verifiable reward~\cite{chen2023textdiffuser} or a learned reward model~\cite{ghosh2023geneval,wu2023human}). Instead of relying solely on pretraining objectives, an external reward can help capture desired structural or perceptual properties to steer generation. Existing approaches primarily focus on \emph{post-hoc alignment}~\cite{liu2025improvevidgen,lu2025rewardforcing}, such as Supervised Fine-Tuning (SFT) or Reinforcement Learning (RL), where model parameters are tilted to reward-favoured distribution. Alternatively, without additional training, reward-based Test-Time Scaling (TTS)~\cite{ma2025inferencescaling,zhang2025inferencescaling} can improve generation by using rewards at inference time to guide search, re-ranking, or iterative refinement among candidate rollouts. Such a paradigm also applies to geometric aspects: recent works~\cite{asim2025met3r,liu2025improvevidgen,kupyn2025epipolar,du2026videogpa} have demonstrated that reward-based post-training (e.g., RL) can effectively improve video generation quality, revealing that the success of reward-based alignment hinges on two key components: (a) a well-designed reward function that faithfully captures the target property (e.g., geometric realism), and (b) a procedure that effectively leverages such reward for alignment, either through parameter-updating post-hoc optimization (e.g., SFT/RL) or inference-time optimization (e.g., TTS).

In this work, unlike prior approaches that rely on handcrafted, rule-based rewards (e.g.,~\cite{gao2025seedance,kupyn2025epipolar,liu2025improvevidgen,lu2025rewardforcing,du2026videogpa}), we propose a geometry-based reward derived from a pretrained geometric foundation model VGGT~\cite{wang2025vggt}. We compute the pointwise reprojection error as a reward signal and additionally leverage the strong representation of VGGT to identify geometrically meaningful regions for guiding sampling. Based on this reward, we curate a diverse prompt suite spanning a wide range of scene types, viewing angles, and camera trajectories, and generate paired videos with contrastive levels of geometric consistency. These pairs form our \textbf{GB3DV-25k} dataset, which provides scalable preference data for geometry-aware alignment. We then apply our reward across multiple downstream settings and model families to demonstrate its effectiveness: (a) \emph{post-hoc alignment}: SFT and preference-based RL of a bidirectional diffusion model (using our dataset); and (b) \emph{test-time scaling}: TTS applied to both a bidirectional diffusion model and a causal autoregressive video model, the latter of which, to the best of our knowledge, has not been previously explored. Within the TTS setting for the causal model, we also explore several reward-guided search strategies enabled by the causal rollout mechanism, showing that autoregressive generation supports more structured inference-time exploration, yielding improved scaling behavior.

Our contributions can be summarized as follows:
\begin{itemize}
    \item We propose a geometry-based reward that measures multi-view consistency via cross-frame pointwise reprojection error using a pretrained geometric foundation model, together with a geometry-aware sampling strategy that focuses evaluation on reliable, semantically meaningful regions.
    \item Leveraging this reward, we curate \textbf{GB3DV-25k}, a preference dataset of 25{,}600 geometry-ranked video pairs spanning diverse scenes and camera motions, and show that SFT and RL improve the geometric consistency of bidirectional video diffusion models.
    \item We study reward-guided TTS for video generation, which pioneers TTS for streaming (causal autoregressive) video models. We propose a general TTS framework and instantiate three efficient search strategies, i.e. \emph{Search on Start}, \emph{Search on Path}, and \emph{Beam Search}, that use our reward as a path verifier to improve geometric realism without retraining.
\end{itemize}

\section{Related Works}
\label{sec:related}

\subsection{Video Diffusion Models}
Recent video diffusion models (VDMs) are predominantly built upon Diffusion Transformer (DiT) architectures~\cite{peebles2023dit} operating in compressed spatiotemporal latent spaces. They follow a paradigm that first encode video through a 3D variational autoencoder (VAE) that applies joint spatial and temporal compression, then denoise the resulting latent representation with a transformer backbone trained under flow matching~\cite{lipman2023fm} or rectified flow objectives~\cite{liu2022rectifiedflow}. According to the token processing approach, video generation models can be further catogerized into \emph{bidirectional} and \emph{causal autoregressive} ones. Bidirectional models~\cite{brooks2024sora,googledeepmind2025veo3,klingteam2025kling,gao2025seedance,yang2025cogvideox,wan2025wan,kong2025hunyuanvideo,polyak2025moviegen} attend to all space-time tokens jointly via full 3D spatiotemporal attention~\cite{ho2022vdm,Gupta2023PhotorealisticVG,brooks2024sora,yang2025cogvideox}, yielding high visual fidelity and strong long-range temporal coherence. However, this design requires denoising all frames simultaneously, leading to attention cost that scales quadratically with the total token count and latency proportional to video length. Causal autoregressive models~\cite{Chen2024DiffusionFN,huang2025selfforcing,yin2025causvid,zhu2026causalforcing} instead generate frames sequentially with KV caching, trading full attention for low-latency streaming. Our test-time scaling strategy is applied on top of causal models in this family, exploiting their sequential generation structure to enable structured inference-time search without any retraining.

\subsection{Reward-Tilted Generation}
Reward-based methods steer generative models by transferring the reward information into the generative distribution through three paradigms. The first is \textit{gradient-based reward fine-tuning}, where model weights are updated to maximize a reward signal, either via supervised fine-tuning (SFT) on high-reward samples~\cite{lee2023alignt2i} or via direct backpropagation through differentiable rewards~\cite{clark2024draft, liu2025improvevidgen}. The second is \textit{particle-based scaling methods}, which search over a discrete set of generation candidates, via Best-of-$N$ sampling (BoN)~\cite{verdun2025bon} or Test-Time Scaling (TTS)~\cite{ma2025inferencescaling,zhang2025inferencescaling}, and select the highest-reward output without updating model parameters. The third is \textit{reinforcement learning}, where policy optimization objectives such as RLHF~\cite{christiano2017rlhf,ouyang2022instructgpt}, DPO~\cite{rafailov2024dpo}, and GRPO~\cite{shao2024deepseekmath} are used to align the model with a reward signal. Recent works have explored distilling geometric priors into video diffusion models via RL-based alignment. Epipolar-DPO~\cite{kupyn2025epipolar} uses Sampson epipolar distance as a geometry-aware DPO preference signal, while concurrent work VideoGPA~\cite{du2026videogpa} extends this to scene-level by reconstructing videos with VGGT~\cite{wang2025vggt} and measuring pixel-level discrepancy via re-rendering. Additional related work is discussed in the supplementary material.
\section{Preliminaries}

\subsection{Video Generation via Flow Matching}
Modern video diffusion models are increasingly built upon \textit{rectified flow}~\cite{liu2022rectifiedflow}, a framework that defines a straight-line transport between a Gaussian noise distribution $p_1 = \mathcal{N}(\mathbf{0}, \mathbf{I})$ and the data distribution $p_0$. Given a clean video sample $x_0 \sim p_0$ and noise $\epsilon \sim \mathcal{N}(\mathbf{0}, \mathbf{I})$, the noisy latent at time $t \in [0, 1]$ is defined by the linear interpolation: $x_t = (1 - t)\, x_0 + t\, \epsilon.$

A neural network $v_\theta(x_t, t)$ is trained to predict the velocity field that transports $x_t$ back toward the data manifold by minimizing the flow-matching objective: $\mathcal{L}_{\text{FM}} = \mathbb{E}_{t, x_0, \epsilon}\left[\left\| v_\theta(x_t, t) - (x_0 - \epsilon) \right\|^2\right].$ At inference time, a clean sample is recovered by numerically integrating the learned velocity field from $t=1$ to $t=0$ via an ODE solver. This formulation underlies several state-of-the-art open-source video generators and serves as the backbone for the preference alignment procedure described in Sec.~\ref{sec:alignment}.

\subsection{Camera Re-Projection}
We briefly review the standard pinhole camera model that underpins our geometric reward computation. Each frame $i$ in a video sequence is associated with a camera intrinsic matrix $\mathbf{K}_i \in \mathbb{R}^{3 \times 3}$ and an extrinsic transformation $[\mathbf{R}_i \mid \mathbf{t}_i] \in \mathbb{R}^{3 \times 4}$, where $\mathbf{R}_i$ and $\mathbf{t}_i$ denote the rotation and translation from world coordinates to the camera frame. In our work, these camera parameters and per-frame depth maps are estimated in a feed-forward manner by geometric foundation models~\cite{wang2024dust3r,leroy2024mast3r,lin2025depth3,wang2025vggt}, which jointly predicts camera intrinsics, extrinsics, and dense depth from a sequence of images without recursive optimization.

Given a 2D pixel location $(u_k, v_k)$ in frame $i$ with associated depth $d_k$, the corresponding 3D point in world coordinates is obtained via back-projection:

\begin{equation}
\label{eq:unproject}
\mathbf{X}_k^w = \mathbf{R}_i^{-1} \left( d_k \mathbf{K}_i^{-1} \begin{bmatrix} u_k, v_k, 1 \end{bmatrix}^{T} - \mathbf{t}_i \right),
\end{equation}
where $\mathbf{X}_k^w = (X_k, Y_k, Z_k)^T$ denotes the world-space coordinate. To reproject this 3D point into a target frame $j$ with camera parameters $(\mathbf{K}_j, \mathbf{R}_j, \mathbf{t}_j)$, we transform it into the target camera space and apply the perspective projection:

\begin{equation}
\label{eq:reproject}
\mathbf{X}_k^c = \mathbf{R}_j \mathbf{X}_k^w + \mathbf{t}_j, \qquad
\begin{bmatrix} \hat{u}_k^{(j)}, \hat{v}_k^{(j)}, 1 \end{bmatrix}^{T} = \frac{1}{Z_k^c}\, \mathbf{K}_j\, \mathbf{X}_k^c,
\end{equation}
where $\mathbf{X}_k^c = (X_k^c, Y_k^c, Z_k^c)^T$ is the camera-space coordinate and $Z_k^c$ is the corresponding depth in frame $j$. The reprojected pixel $\hat{\mathbf{p}}_k^{(j)} = (\hat{u}_k^{(j)}, \hat{v}_k^{(j)})^T$ serves as a prediction of where point $k$ should appear in frame $j$, and its deviation from the tracker-predicted correspondence constitutes our reward signal (Sec.~\ref{sec:reproj}).

\begin{figure}[t]
\centering
\includegraphics[width=\textwidth]{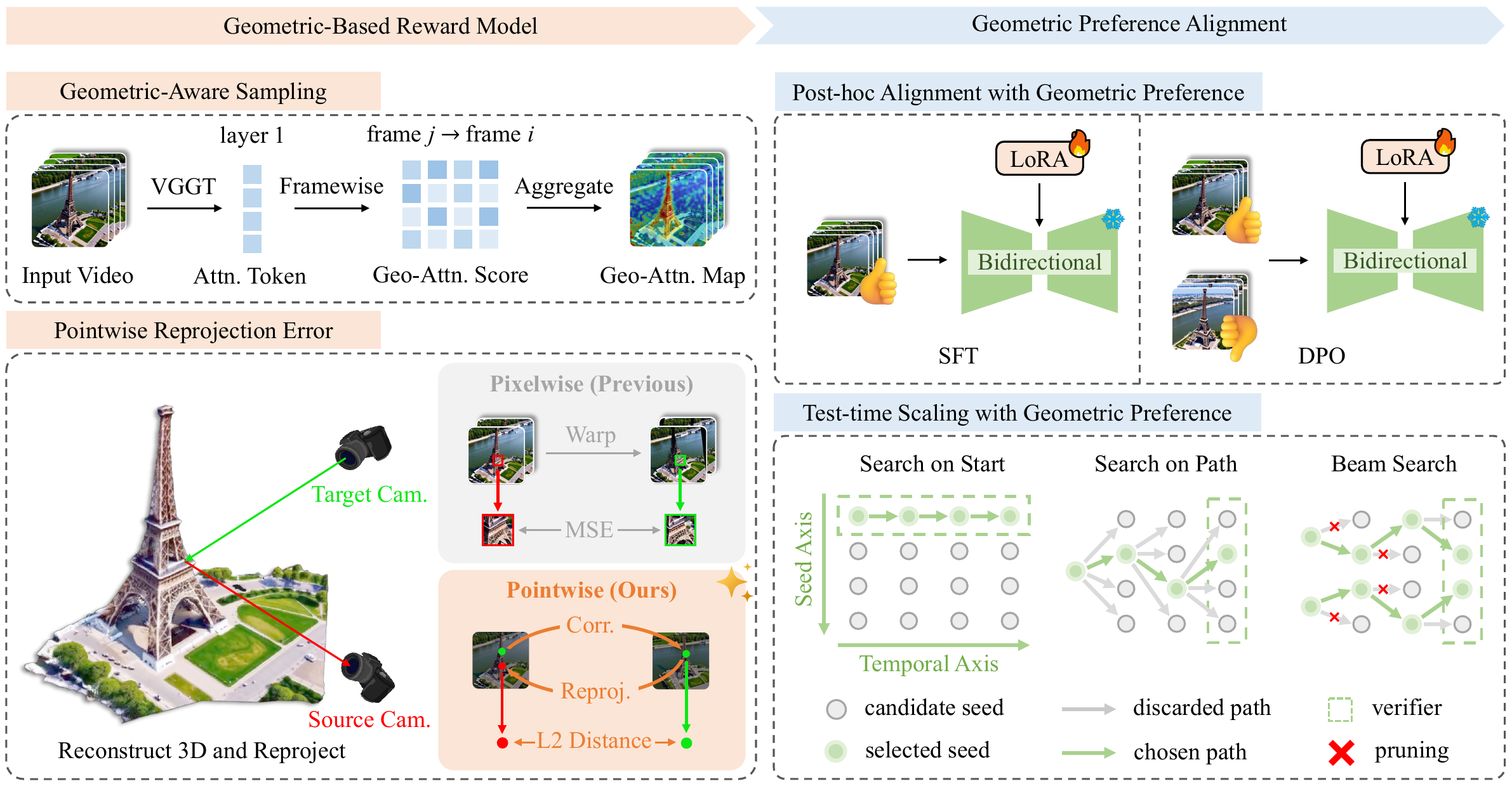}
\caption{\textbf{Overview of our framework}, consisting of two components: (a) \textbf{Geometric-Based Reward Model}: a geometry-aware sampling (GAS) module leverages global attention of VGGT to identify salient patches and computes cross-frame pointwise reprojection error; (b) \textbf{Geometric Preference Alignment}: the model is aligned via SFT~\cite{lee2023alignt2i} and DPO~\cite{liu2025improvevidgen} on a bidirectional model, or test-time scaling (TTS) with our reward as a path verifier on a causal model\cite{zhu2026causalforcing}.}
\label{fig:overview}
\end{figure}

\section{Geometry-Based Reward}

\begin{figure}[t]
\centering
\textit{\footnotesize ``Wide shot of the \textcolor{red}{Brandenburg Gate} under an overcast sky. Camera dollies in ...''}

\def\labelwd{0.03\textwidth}
\def\imgwd{0.25\linewidth}

\begin{minipage}[c]{\labelwd}
    \rotatebox{90}{%
        \parbox{1cm}{\centering\scriptsize Input Video}%
    }
\end{minipage}%
\hspace{1mm}
\begin{minipage}[c]{0.95\textwidth}
    \includegraphics[width=\imgwd]{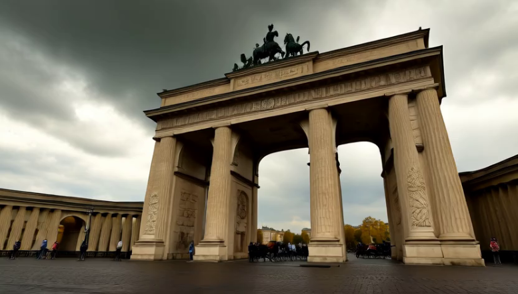}\hfill%
    \includegraphics[width=\imgwd]{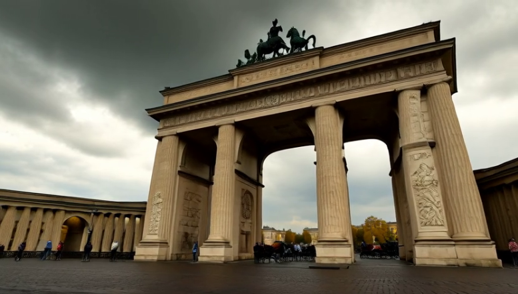}\hfill%
    \includegraphics[width=\imgwd]{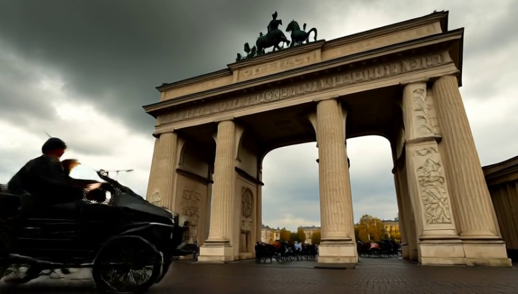}\hfill%
    \includegraphics[width=\imgwd]{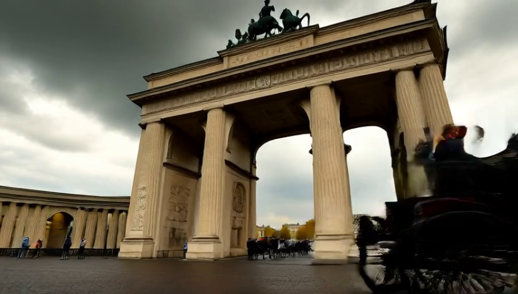}%
\end{minipage}

\begin{minipage}[c]{\labelwd}
    \rotatebox{90}{%
        \parbox{1cm}{\centering\scriptsize Attention Maps}%
    }
\end{minipage}%
\hspace{1mm}
\begin{minipage}[c]{0.95\textwidth}
    \includegraphics[width=\imgwd]{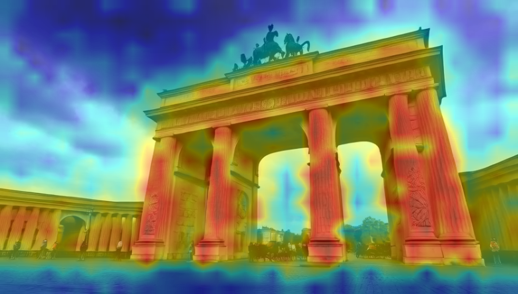}\hfill%
    \includegraphics[width=\imgwd]{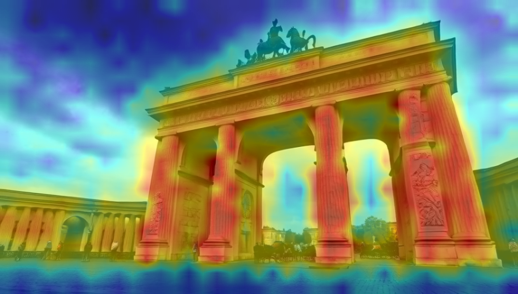}\hfill%
    \includegraphics[width=\imgwd]{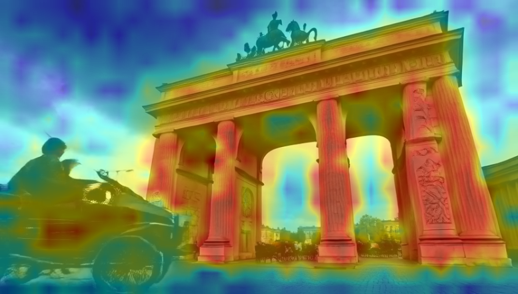}\hfill%
    \includegraphics[width=\imgwd]{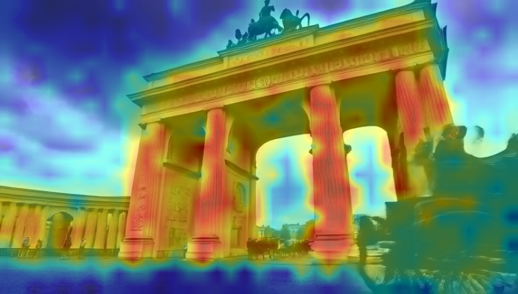}%
\end{minipage}

\begin{minipage}[c]{\labelwd}
    \rotatebox{90}{%
        \parbox{1cm}{\centering\scriptsize Sampled Points}%
    }
\end{minipage}%
\hspace{1mm}
\begin{minipage}[c]{0.95\textwidth}
    \includegraphics[width=\imgwd]{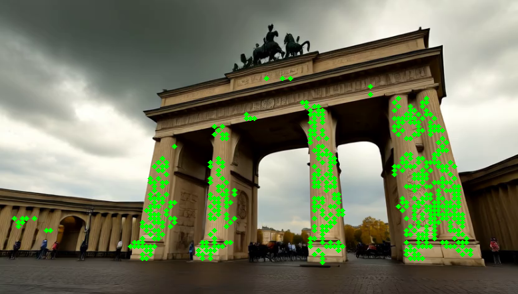}\hfill%
    \includegraphics[width=\imgwd]{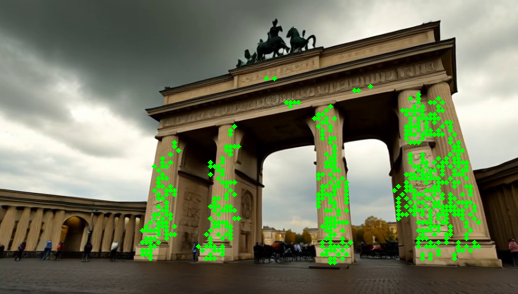}\hfill%
    \includegraphics[width=\imgwd]{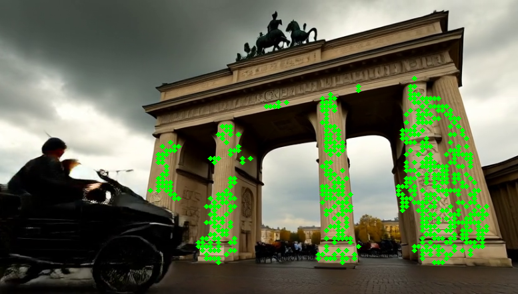}\hfill%
    \includegraphics[width=\imgwd]{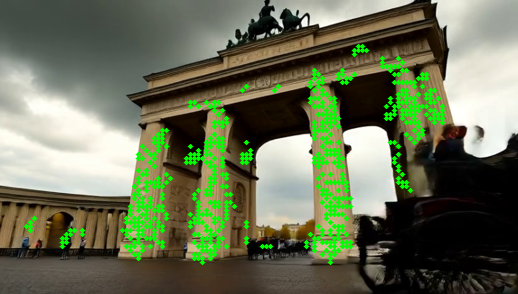}%
\end{minipage}

\caption{\textbf{Visualization of Geometry-Aware Sampling}, which shows that the global attention of VGGT naturally captures the background geometry. We select top-$\tau$ percentage of attention-emphasized patches and sample at the center of each patch.}
\label{fig:gas}
\end{figure}

To align video generative models with geometric preference, we introduce Geometry-Based Reward, which takes advantage of an existing geometric foundation model (VGGT~\cite{wang2025vggt}) to conduct dense reconstruction and evaluate three-dimensional consistency by point-wise reprojection error. While dense reconstruction provides a geometry-grounded representation of the video, simply computing pixel-space error via warping frame to frame would introduce significant interference. To remedy the issue, as shown in Fig~\ref{fig:overview}-(left), we propose a \emph{geometry-aware sampling strategy} that filters out distractors and \emph{computes reprojection error} in a pointwise manner to capture three-dimensional inconsistencies. 

\subsection{Geometry-Aware Sampling}
Previous studies~\cite{han2025emergent,hu2025vggt4d,bratulic2025geometric,cao2026vggtdet} show that the global attention layers of VGGT excavate global information across input frames. For geometric context, we find that the shallow global-attention layers constantly emphasize geometrically meaningful areas (Fig.~\ref{fig:gas}, middle row), as they process more geometric properties of the scene. We leverage this to select geometrically critical areas and filter out irrelevant regions (e.g., sky or ground).

Given a video sequence $\mathcal{I}=\{I_i\}_{i=0}^N$, we pass it through alternating-attention layers~\cite{wang2025vggt} and extract the shallow layer to compute the geometric-attention score. Anchoring frame $i$ as the query and the others $j$ as keys, we compute the scaled dot-product attention of the query tokens $\mathbf{Q}_i$ to key tokens $\mathbf{K}_j$:

\begin{equation}
\label{eq:gas}
\mathbf{A}_i = \frac{1}{N-1}\sum_{j \ne i}^{N-1} \text{softmax}\left(\frac{\mathbf{Q}_i \mathbf{K}_j^T}{\sqrt{d}}\right),
\end{equation}
where $d$ is the attention dimension. These token-level scores are averaged across heads and summed over all frames $j \neq i$, then up-sampled to full resolution $(H, W)$ via bilinear interpolation and normalized to $[0,1]$ to produce a geometric attention heatmap $\mathbf{M}_i$.

We partition each frame into non-overlapping $p \times p$ patches, with each patch's value being the mean of $\mathbf{M}_i$ within it, and select the top $\tau\%$ by attention value. For each selected patch, we take its center pixel as a sampling point (Fig.~\ref{fig:gas}, bottom row), yielding 2D point locations $\mathcal{P}^{(i)}_{k} = \{(u^{(i)}_k, v^{(i)}_k)\}_{k=1}^{K^{(i)}}$ per frame $i$, where $K^{(i)}$ is the number of retained patches. By default we set $p=4$ and $\tau=20\%$ in all experiments.

\subsection{Pointwise Reprojection Error}
\label{sec:reproj}

After identifying geometrically critical regions in each frame, we can establish point correspondences across frames and compute pointwise reprojection errors to quantify geometric consistency.

\subsubsection{Point Tracking and Correspondence.}
For each reference frame $i$ with sampled points $\mathcal{P}_i$, we employ a tracking module to establish correspondences by querying the tracker with points $\mathcal{P}_i$ against all other frames $j$. This yields tracked positions $\mathbf{p}_k^{(j)} = (u_k^{(j)}, v_k^{(j)})^T$ for each point $k$ across all frames $j \in \{0, \ldots, N-1\}$, along with confidence scores $c_k^{(j)} \in [0, 1]$.

\subsubsection{Unprojection to 3D Space.}
For each query point $\mathbf{p}_k$ in reference frame $i$, we retrieve its depth value $d_k = D_i[v_k, u_k]$ from the VGGT-predicted depth map $D_i$, together with the corresponding camera parameters $(\mathbf{K}_i, \mathbf{R}_i, \mathbf{t}_i)$. The 3D world coordinate $\mathbf{X}_k^w$ is then recovered via back-projection as defined in Eq.~\ref{eq:unproject}.

\subsubsection{Reprojection and Error Computation.}
To validate geometric consistency, we reproject each 3D point $\mathbf{X}_k^w$ into target frame $j$ using Eq.~\ref{eq:reproject}, yielding the geometry-grounded estimate $\hat{\mathbf{p}}_k^{(i \to j)}$. We apply validity filtering of sampled points to ensure robust error computation. The final reprojection error metric is the mean L2 distance over all valid point-frame pairs:

\begin{equation}
\mathcal{E}_{\text{reproj}} = \frac{1}{|\mathcal{V}|} \sum_{(k,i,j) \in \mathcal{V}} \|\hat{\mathbf{p}}_k^{(i \to j)} - \mathbf{p}_k^{(j)}\|_2,
\end{equation}
where $\mathcal{V}$ denotes the set of valid point-frame pairs. A lower reprojection error indicates stronger geometric consistency between the tracker's correspondences and the foundation model's predicted geometry.

\section{Geometry-Guided Preference Alignment}
\label{sec:alignment}

Once a reliable reward model is available, it can be used to influence video generation in several ways: (a) \emph{online optimization}, where the model is updated using reinforcement learning with reward feedback, such as RLHF~\cite{christiano2017rlhf,ouyang2022instructgpt} and GRPO~\cite{shao2024deepseekmath}; (b) \emph{offline optimization}, which includes post-hoc preference learning methods such as DPO~\cite{rafailov2024dpo} or supervised fine-tuning~\cite{lee2023alignt2i} on offline generated data; and (c) \emph{test-time optimization}, where the reward guides search, ranking, or iterative refinement of candidate generations without updating model parameters. Due to limitations in model size and computational speed, most video generation methods adopt \emph{offline optimization}~\cite{kupyn2025epipolar,du2026videogpa}, particularly SFT or DPO, to incorporate reward knowledge into the generation process.

\subsection{Post-hoc Preference Alignment}

\subsubsection{Preference Data Construction.}
For offline optimization, a dataset must first be constructed using the reward signal to selectively curate or annotate samples. We construct our \textbf{GB3DV-25k} dataset as follows:
Given a fixed set of random seeds $\{s_i\}_{i=0}^N$ and condition vector $c$, we generate a video set via video diffusion $x_0 = G_{\theta}(z_i, c), z_i \sim \mathcal{N}(\mathbf{0}, \mathbf{I})$ where $z_i$ is the Gaussian noise generated with seed $s_i$. We then leverage the proposed geometry-based reward as evaluator $r_{geo}$ to assess the videos. To magnify the geometric difference signal across videos, we pick the best and worst samples for each prompt and construct the $(x_0^w,x_0^l)$ pair as our preference data.

\subsubsection{Supervised Fine-Tuning.}
The most direct way to leverage a reward signal is to perform supervised fine-tuning (SFT) in the reward-selected dataset $\mathcal{D^*}$. In this setting, we adopt LoRA~\cite{hu2021lora}, a parameter-efficient fine-tuning (PEFT) approach, on the video generator and train the model $v_\theta$ using the same diffusion/flow matching objective:

\begin{equation}
\mathcal{L}_{\text{FM}} = \mathbb{E}_{t, x_0 \in \mathcal{D^*}, \epsilon}\left[\left\| v_\theta(x_t, t) - (x_0 - \epsilon) \right\|^2\right].
\end{equation}

\subsubsection{DPO with Geometric Preference.}
Direct Preference Optimization (DPO) is an offline reinforcement learning method that aligns the policy using pairwise preferences under the Bradley--Terry model. It leverages preference comparisons to directly optimize the policy without requiring a globally normalized reward function. Given preference pairs $(x_0^w, x_0^l)$, the geometric preference DPO optimizes the parameters $\theta$ via:

\begin{equation}
\max_{\theta} \mathbb{E}_{(x_0^w, x_0^l) \sim \mathcal{D}} \left[ \log \sigma \left( \beta \log \frac{p_\theta(x_0^w \mid c)}{p_{\text{ref}}(x_0^w \mid c)} - \beta \log \frac{p_\theta(x_0^l \mid c)}{p_{\text{ref}}(x_0^l \mid c)} \right) \right],
\end{equation}
where $\beta$ is the coefficient that controls the deviation from the reference policy $p_{\text{ref}}$, and $\mathcal{D}$ denotes the preference dataset containing $x_0^w \succ x_0^l$ ranked by $r_{geo}$.

In our work, we adopt the formulation of Flow-DPO~\cite{liu2025improvevidgen}, which reformulates the objective for rectified flow models:

\begin{equation}
\begin{aligned}
\mathcal{L}=-\mathbb{E}[\log \text { sigmoid }(- & \frac{\beta_{t}}{2}\left(\left\|v^{w}-v_{\theta}\left(\mathbf{x}_{t}^{w}, t\right)\right\|^{2}-\left\|v^{w}-v_{\mathrm{ref}}\left(\mathbf{x}_{t}^{w}, t\right)\right\|^{2}\right. \\
& \left.\left.\left.-\left(\left\|v^{l}-v_{\theta}\left(\mathbf{x}_{t}^{l}, t\right)\right\|^{2}-\left\|v^{l}-v_{\mathrm{ref}}\left(\mathbf{x}_{t}^{l}, t\right)\right\|^{2}\right)\right)\right)\right],
\end{aligned}
\end{equation}
where $v_{ref}$ and $v_\theta$ are the velocities predicted by the model, $\beta_t = \beta(1-t^2)$ is the training weight that depends on the noise level, and $x_t^{*} = (1-t)x_0^{*}+t\epsilon$ is the noisy latent of the preference pair.

To prevent mode collapse, we introduce two auxiliary loss terms: a first-order term that penalizes static motion and a second-order term that encourages overall smoothness, applied with opposite signs:

\begin{equation}
\label{eq:aux}
\mathcal{L}_{\text{aux}} = -\mathbb{E}\bigl[\mathrm{Var}_t(\hat{x}_0)\bigr]
+ \gamma\cdot\mathbb{E}\bigl[\|\Delta^2_t\, \hat{x}_0\|^2\bigr],
\end{equation}
where $\hat{x}_0$ is the reconstructed clean sample, $\Delta^2_t\,\hat{x}_0$ denotes the second-order temporal difference of $\hat{x}_0$, and $\gamma> 0$ is the weighting coefficient.
The final loss of DPO is: $\mathcal{L}_{total}=\mathcal{L} + \lambda \mathcal{L}_{aux}$, where $\lambda$ controls the strength of the penalty.

\subsection{TTS with Geometric Preference}
In contrast to most \textit{post-hoc} alignment methods that require constructing a dataset beforehand, Test-Time Scaling (TTS) optimizes the reward signal directly at inference time. It aims to identify an optimal sample using a verifier $\mathcal{V}$ together with a search procedure $f$:

\begin{equation}
    f : \mathcal{V} \times G_\theta \times \{\mathbb{R}^{N \times H \times W \times C} \times \mathbb{R}^d\}^N \rightarrow \mathbb{R}^{N \times H \times W \times C},
\end{equation}
where $G_\theta$ is the pretrained generator and $\mathcal{V}$ evaluates the quality of candidates. 

For bidirectional models, which denoise all spatial-temporal tokens jointly, TTS reduces to a Best-of-$N$ protocol similar to most image generation TTS scenarios~\cite{ma2025inferencescaling,zhang2025inferencescaling}: $N$ candidates are generated in parallel, and the highest-rewarded sample is returned. Causal autoregressive models, by contrast, generate frames sequentially, where each frame $x_t$ is conditioned on the previously generated prefix $x_{<t}$. This temporal Markov structure exposes a richer search space: one can intervene at each step to prune low-quality paths and explore diverse trajectories, going beyond the simple Best-of-$N$ selection that only retains a narrow search space.

To exploit this richer search space, we instantiate TTS on a causal autoregressive model~\cite{zhu2026causalforcing} and reformulate the problem as searching for a generation path in a space spanned by the seed axis and temporal axis, with the goal of finding the highest-rewarded sample.  Applying brute-force search guarantees an optimal solution, but it yields intractable $\mathcal{O}(S^N)$ complexity for a video clip of $N$ frames with a seed search range of size $S$. To address this, we propose three search algorithms (shown in Fig.~\ref{fig:overview}, right bottom) that all achieve a reasonable complexity but with different dynamics:

\subsubsection{Search on Start (SoS).}
The algorithm operates along the seed axis. Given a fixed set of $S$ candidate seeds $\{s_1, s_2, \ldots, s_S\}$, the algorithm performs a complete forward pass for each seed independently and evaluates the resulting video clip using the reward model $r_{geo}$. The seed yielding the highest reward is selected, and its corresponding output clip is returned:

\begin{equation}
\label{eq:sos}
    s_{best} = \arg\max_{s_{i}} \, r_{geo}\!\left[G_\theta(z_i, c)\right], \quad i = 1, \ldots, S.
\end{equation}

\subsubsection{Search on Path (SoP).}
The algorithm proceeds along the temporal axis, dynamically selecting from a set of seed candidates $S$ at each time step $t$. When generating the next frame, it iterates over $S$ and selects the optimal seed $s^{t}_{best}$ that yields the highest reward. The reward model evaluates frames within a sliding context window $\mathcal{W}$ spanning the preceding $w$ frames, including the current one. The seed path that achieves the highest cumulative reward is then returned:

\begin{equation}
\label{eq:sop}
    s^{t}_{best} = \arg\max_{s^{t}_{i}} \, r_{geo}\!\left[\mathcal{W} \cup \left\{G_\theta(z^{t}_{i}, c)\right\}\right], \quad i = 1, \ldots, S.
\end{equation}

\subsubsection{Beam Search (BS).}
The algorithm generalizes both SoS and SoP by maintaining $K$ candidate paths throughout generation. At each time step, it evaluates $K \times S$ child nodes spawned from all current paths and retains only the top-$K$ nodes, pruning the rest. The path achieving the highest cumulative reward is returned. The objective follows the same formulation as Eq.~\ref{eq:sop}, with the distinction that $K$ optimal seeds $s^{t}_{best}$ are retained at every step rather than one.

It's noteworthy that the previous two search plans can be viewed as special configurations of BS: SoS corresponds to $(K, 1)$ and SoP corresponds to $(1, S)$, with complexities of $\mathcal{O}(KN)$ and $\mathcal{O}(SN)$. The beam search algorithm supports flexible $(K, S)$ configurations, yielding a general complexity of $\mathcal{O}(KSN)$.

\section{Experiments}

\subsection{Implementation Details}

\subsubsection{Dataset Curation.}
Preference alignment requires training pairs with notable differences in geometric consistency. To ensure sufficient diversity, we generate 10 samples per prompt using bidirectional CausVid~\cite{yin2025causvid}. To construct the prompt suite, we source scene references from two representative datasets: RealEstate10k~\cite{zhou2018stereomagnification} for indoor scenes and GLDv2~\cite{weyand2020gldv2} for outdoor scenes. We employ Qwen3-VL~\cite{qwen3vl} to generate detailed scene captions, explicitly instructing the model to describe both static and dynamic objects, diverse camera movements, and varying shooting angles. The resulting prompt suite comprises 2,560 entries, from which CausVid generates a total of 25,600 video clips, constituting our curated GB3DV-25k dataset.

\subsubsection{Evaluation Metrics.}
For geometric consistency, we adopt two complementary protocols: 3D reconstruction quality for dense evaluation and multi-view consistency metrics for sparse evaluation. For 3D reconstruction quality, we apply VGGT~\cite{wang2025vggt} to uniformly sampled frames to obtain depth maps and camera poses, reproject the recovered 3D point cloud into each target frame, and measure reprojection fidelity via PSNR, SSIM, and LPIPS. For multi-view consistency, we compute three scores on the evaluated videos: epipolar consistency (EPI), reprojection-pixelwise (RPX), and reprojection-pointwise (RPT). For overall video quality, we use VBench~\cite{huang2024vbench}, a comprehensive benchmark covering subject consistency (SC), background consistency (BC), motion smoothness (MS), dynamic degree (DD), aesthetic quality (AQ), and imaging quality (IQ). All VBench scores reported in our tables are normalized using the empirical minimum and maximum values specified in the VBench paper~\cite{huang2024vbench}.

\subsubsection{Baselines.}
We compare our methods against the following baselines:

\noindent\textit{Base Model.}\quad We use the Causvid~\cite{yin2025causvid} for TTS of bidirectional experiments (Sec.~\ref{sec:TTS_Bidirectional}), Causal-Forcing~\cite{zhu2026causalforcing} for TTS of streaming video experiments (Sec.~\ref{sec:TTS_Streaming}), and Wan2.1-T2V-1.3B~\cite{wan2025wan} for post-hoc alignment experiments (Sec.~\ref{sec:TTS_Post-hoc}).

\noindent\textit{Epipolar.}\quad Built directly on the official code of~\cite{kupyn2025epipolar}, assessing geometric consistency via epipolar constraints quantified by the Sampson distance.

\noindent\textit{Reproj-Pix.}\quad A faithful reproduction of VideoGPA~\cite{du2026videogpa}, which employs VGGT to estimate camera geometry and quantifies the reward via pixelwise intensity differences between the source and warped frames.

\subsection{TTS of Bidirectional Video Generation}
\label{sec:TTS_Bidirectional}

\subsubsection{Evaluation Setups.} 
We evaluate the effectiveness of test-time scaling on bidirectional video generation by adopting the best-of-$N$ sampling protocol. During evaluation, each reward model independently selects the highest-scoring video from $N$ generated candidates. For generality, the experiments are conducted on the full-scale GB3DV-25k dataset and $N$ is set to 10.

\subsubsection{Evaluation Results.}
Quantitative results are shown in Tab.~\ref{tab:metric}. We omit each method's score on its own selection metric (marked "-"), as such self-referential entries are trivially optimal and obscure a fair comparison. Our \textit{Reproj-Pts} reward achieves the best results on all three 3D reconstruction metrics and the best EPI and RPX among multi-view metrics, showing that pointwise reprojection error is a more robust and physically grounded signal than pixel-space alternatives. The consistent gain over Reproj-Pix~\cite{du2026videogpa} confirms the benefit of decoupling geometric error from pixel intensity. For overall quality, our method attains the highest total VBench score (84.52\%) while leading on subject consistency and aesthetic quality, indicating that optimizing geometric consistency does not sacrifice perceptual quality.

\begin{table}[t]
\centering
\caption{\textbf{Evaluation of Test-Time Scaling via Best-of-N Sampling}.}
\label{tab:metric}

\setlength{\tabcolsep}{5.5pt}
\renewcommand{\arraystretch}{1.15}

\resizebox{\textwidth}{!}{%
\begin{tabular}{l | ccc ccc | cccccc}
\toprule
\multirow{2}{*}{\tightbold{Method}}
& \multicolumn{6}{c|}{\tightbold{Geometric Consistency Evaluation}} 
& \multicolumn{6}{c}{\tightbold{Overall Video Quality Evaluation (VBench \%)}}\\
\cmidrule(lr){2-7} \cmidrule(lr){8-13}
& \tightbold{PSNR} $\uparrow$ 
& \tightbold{SSIM} $\uparrow$ 
& \tightbold{LPIPS} $\downarrow$ 
& \tightbold{EPI} $\downarrow$ 
& \tightbold{RPX} $\downarrow$ 
& \tightbold{RPT} $\downarrow$ 
& \tightbold{SC} $\uparrow$ 
& \tightbold{BC} $\uparrow$ 
& \tightbold{MS} $\uparrow$ 
& \tightbold{AQ} $\uparrow$ 
& \tightbold{IQ} $\uparrow$ 
& \tightbold{Total} $\uparrow$ \\
\midrule
Baseline~\cite{yin2025causvid} 
& 19.68 & 0.6381 & 0.3604 & 5.553 & 0.9738 & 4.706 
& 94.16 & 92.36 & 97.16 & 57.28 & 75.68 & 83.33 \\

Epipolar~\cite{kupyn2025epipolar} 
& \underline{22.45} & \underline{0.7557} & \underline{0.2432} & - & \underline{0.9546} & \tightbold{2.815}
& \underline{96.22} & \tightbold{94.02} & \tightbold{98.14} & \underline{58.20} & 75.91 & \underline{84.50} \\

Reproj-Pix~\cite{du2026videogpa} 
& 21.07 & 0.7267 & 0.3036 & \underline{4.549} & - & \underline{3.473}
& 95.21 & 92.97 & 97.48 & 58.04 & \tightbold{76.14} & 83.97 \\

\rowcolor{lightgray}
Reproj-Pts \tightbold{(Ours)} 
& \tightbold{22.66} & \tightbold{0.7665} & \tightbold{0.2330} & \tightbold{3.442} & \tightbold{0.9539} & -
& \tightbold{96.39} & \underline{93.95} & \underline{97.96} & \tightbold{58.23} & \underline{76.09} & \tightbold{84.52} \\

\bottomrule
\end{tabular}%
}
\end{table}

\subsection{TTS of Streaming Video Generation}
\label{sec:TTS_Streaming}

\subsubsection{Evaluation Setups.}
We apply test-time scaling to streaming video generation and evaluate the three proposed search algorithms: Search on Start (SoS), Search on Path (SoP), and Beam Search (BS).  We conduct a budget scaling study spanning budgets from 1 to 16 (i.e. budget of 4 means 4 seeds for SoS and SoP, 2 child nodes and top-2 for BS) on a 16-clip subset of GB3DV-25k to investigate the scaling behavior of different searching schemes of TTS.

\subsubsection{Evaluation Results.}
\begin{figure}[t!]
\centering
\includegraphics[width=\textwidth]{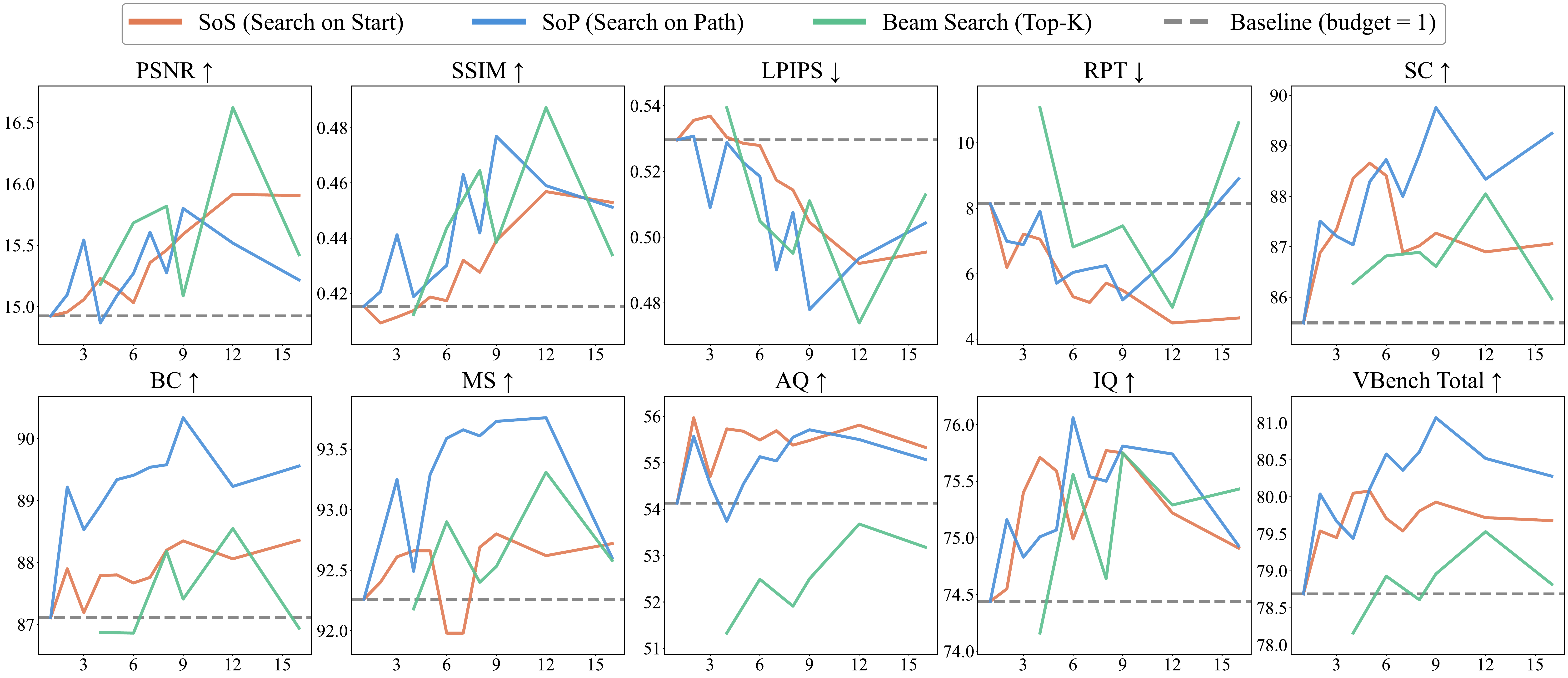}
\caption{\tightbold{Budget Evaluation of TTS}. All three methods—SoS, SoP, and Beam Search—demonstrate scaling tendencies as the search budget increases. Beam Search prevails in 3D metrics, while SoP achieves the best overall performance.}
\label{fig:budget_eval}
\end{figure}

\begin{figure}[t!]
\centering
\def\imgwd{0.25\linewidth}
\begin{minipage}[c]{\textwidth}
    \redbox{\includegraphics[width=\imgwd]{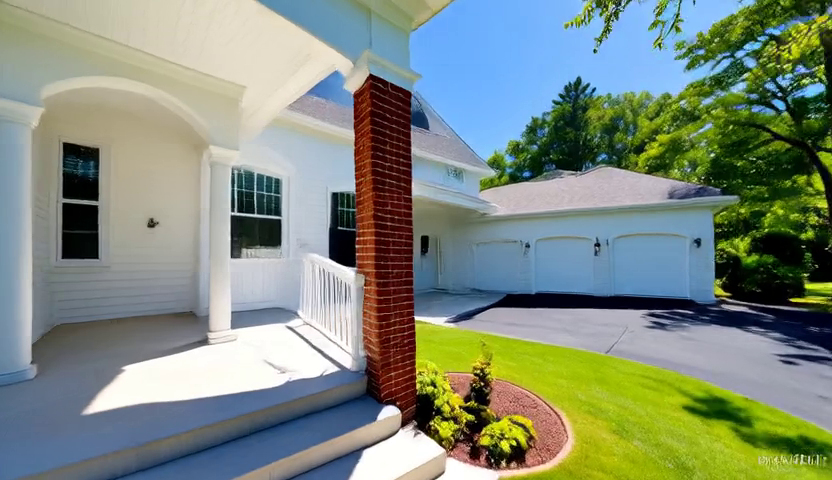}}%
    \includegraphics[width=\imgwd]{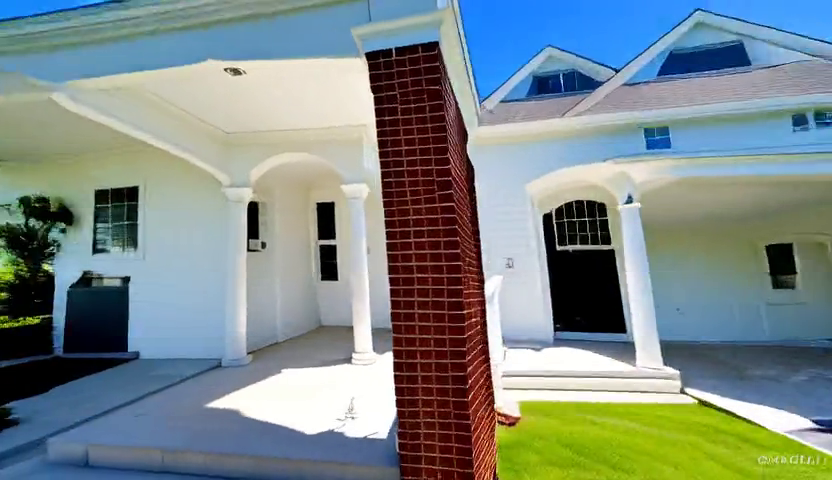}%
    \includegraphics[width=\imgwd]{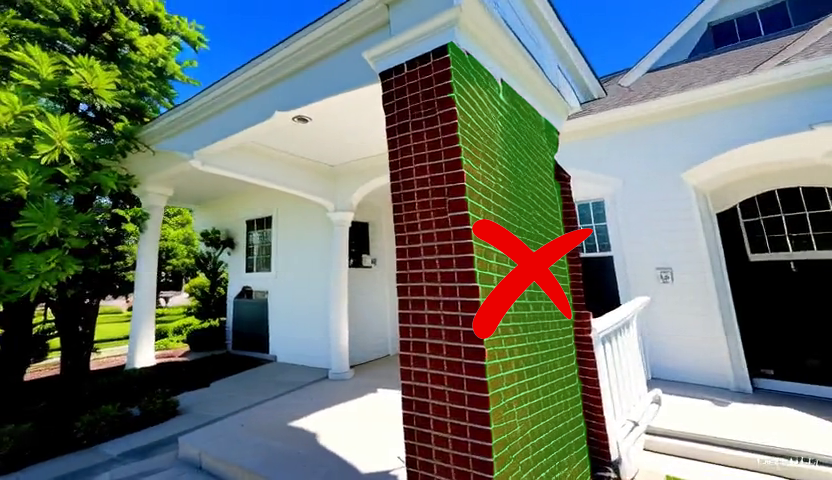}%
    \includegraphics[width=\imgwd]{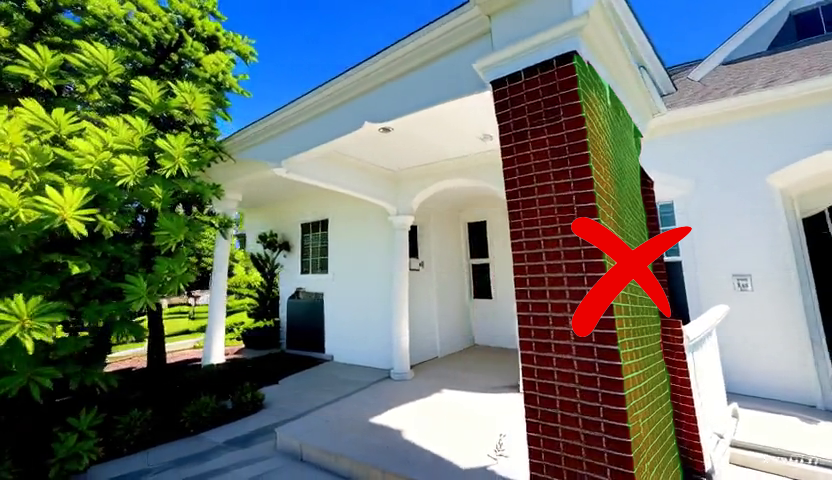}%
\end{minipage}
\begin{minipage}[c]{\textwidth}
    \centering\footnotesize Baseline~\cite{zhu2026causalforcing}
\end{minipage}

\begin{minipage}[c]{\textwidth}
    \redbox{\includegraphics[width=\imgwd]{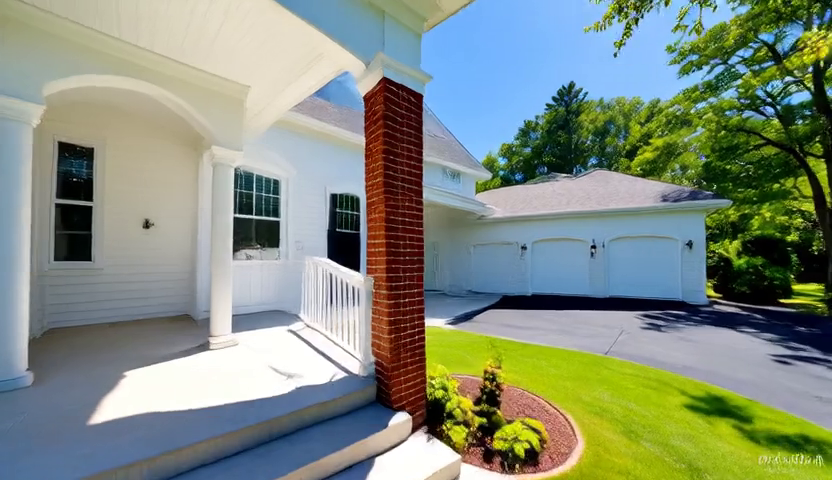}}%
    \includegraphics[width=\imgwd]{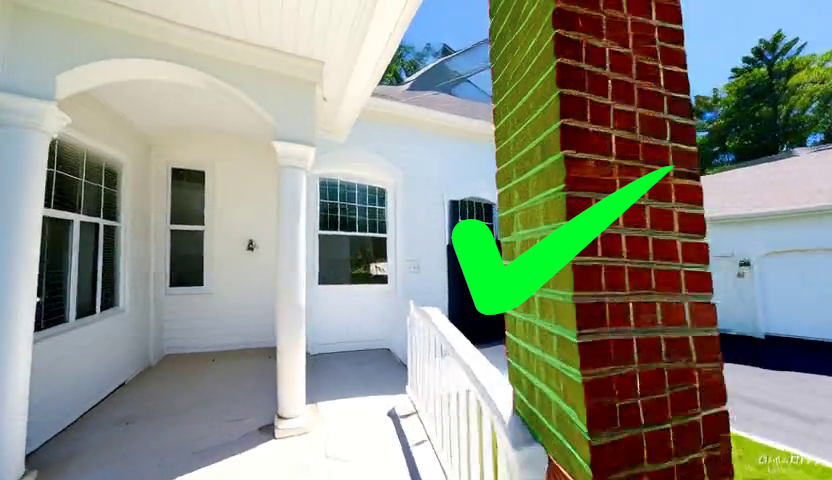}%
    \includegraphics[width=\imgwd]{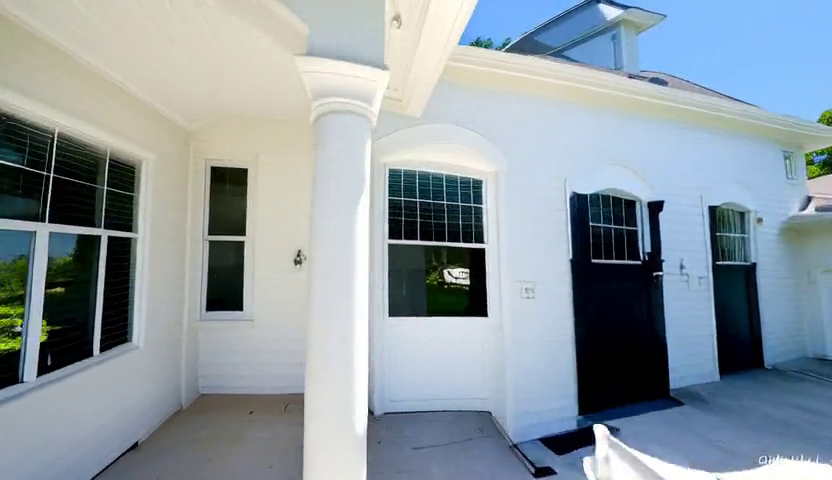}%
    \includegraphics[width=\imgwd]{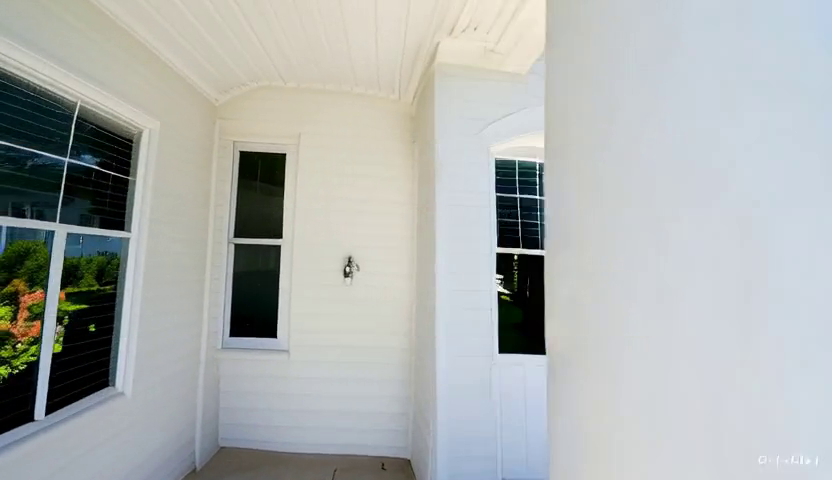}%
\end{minipage}
\begin{minipage}[c]{\textwidth}
    \centering\footnotesize Baseline + Search on Path
\end{minipage}

\begin{minipage}[c]{\textwidth}
    \includegraphics[width=\imgwd]{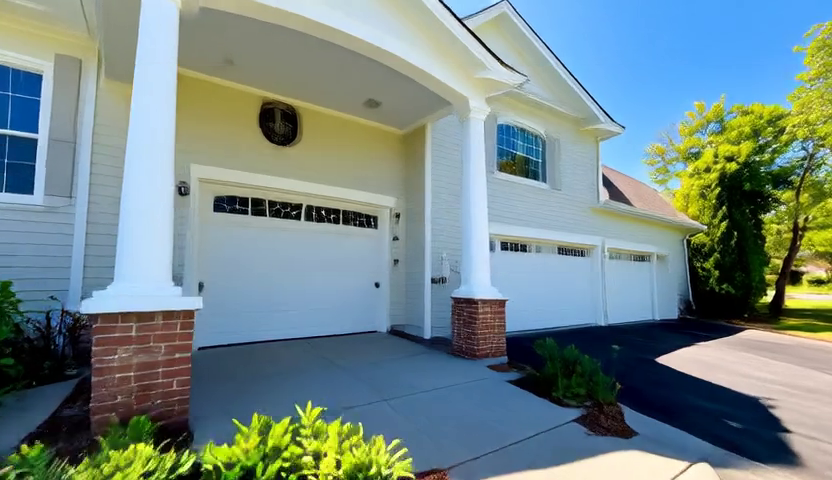}%
    \includegraphics[width=\imgwd]{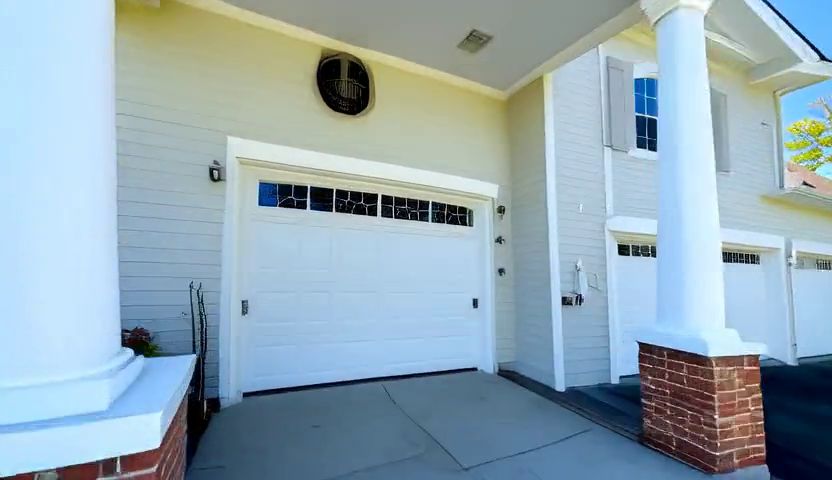}%
    \includegraphics[width=\imgwd]{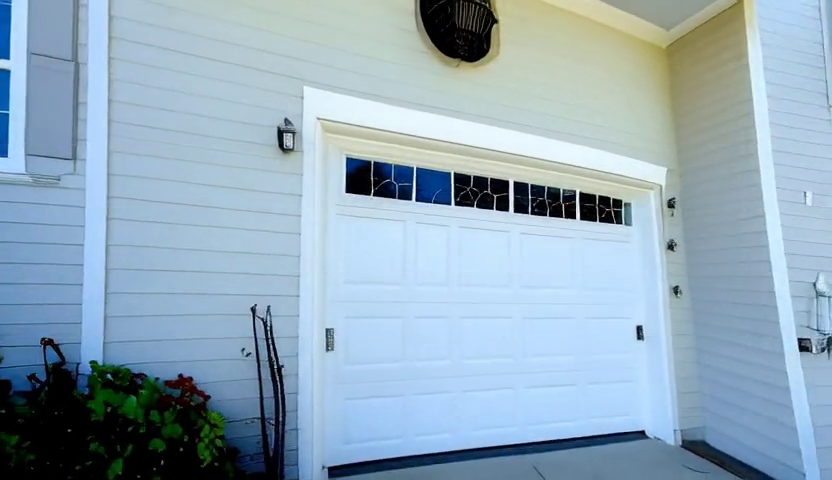}%
    \includegraphics[width=\imgwd]{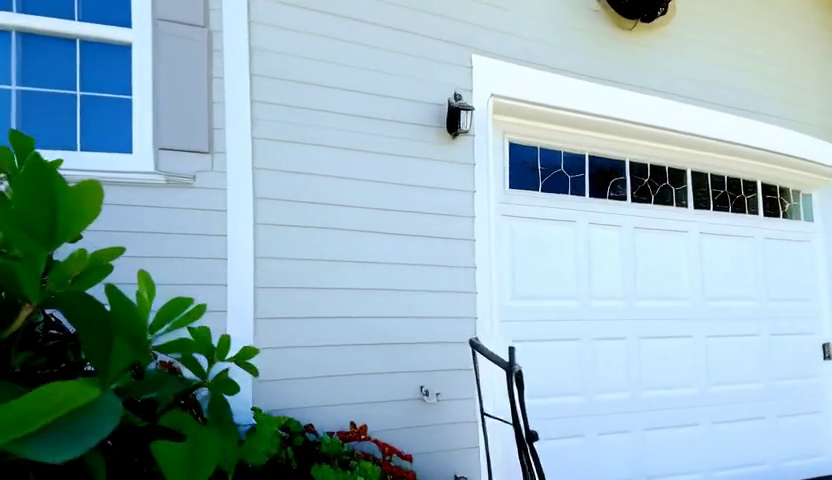}%
\end{minipage}
\begin{minipage}[c]{\textwidth}
    \centering\footnotesize Baseline + Search on Start
\end{minipage}

\begin{minipage}[c]{\textwidth}
    \includegraphics[width=\imgwd]{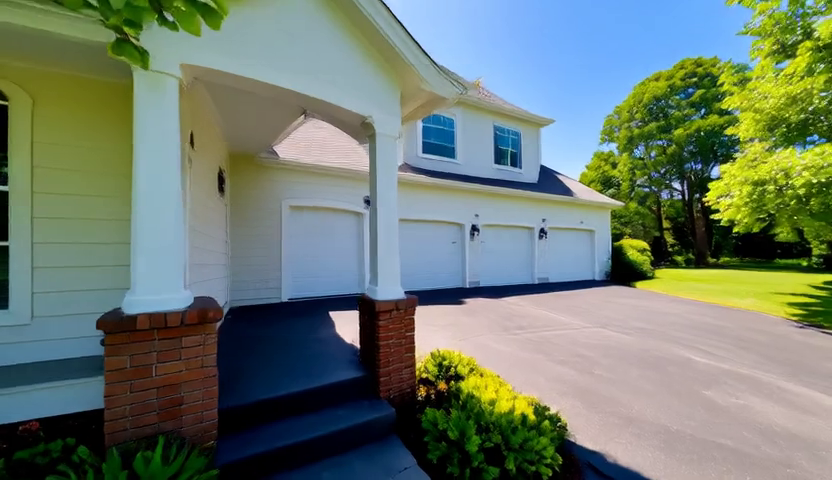}%
    \includegraphics[width=\imgwd]{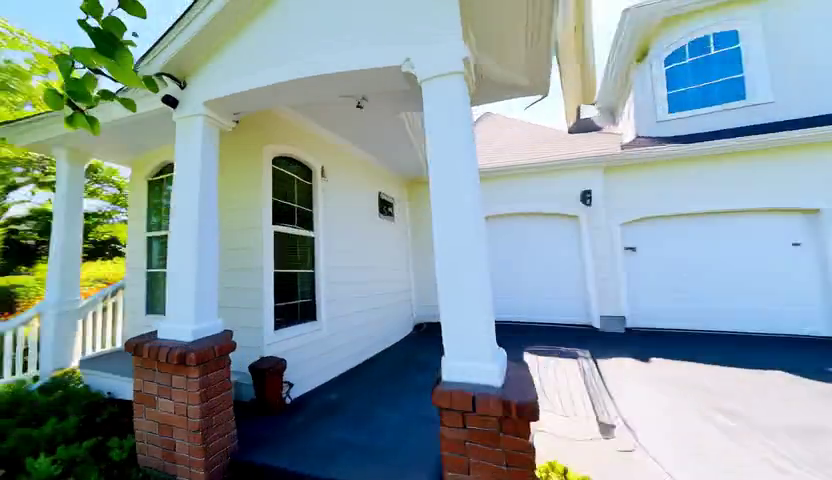}%
    \includegraphics[width=\imgwd]{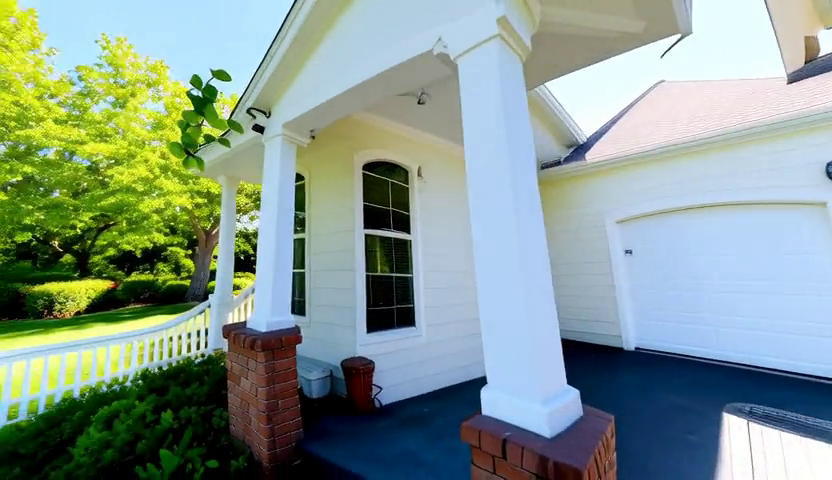}%
    \includegraphics[width=\imgwd]{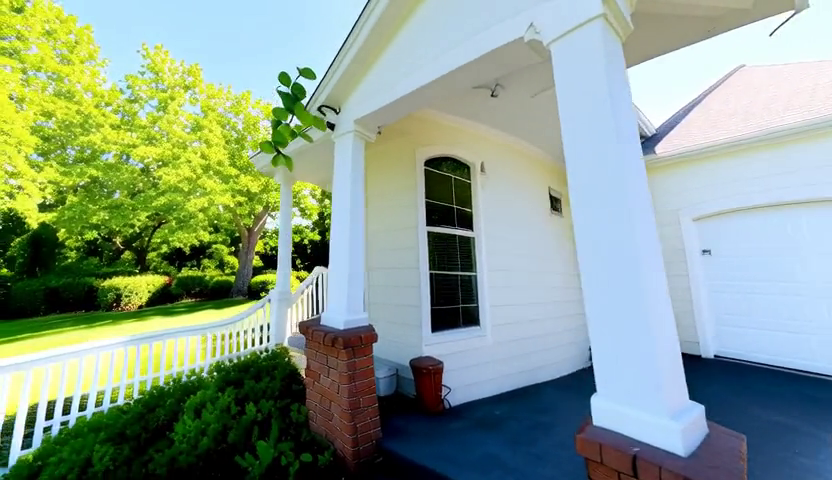}%
\end{minipage}
\begin{minipage}[c]{\textwidth}
    \centering\footnotesize Baseline + Beam Search
\end{minipage}

\caption{\textbf{Qualitative Results of Test-Time Scaling.} The baseline exhibits geometric artifacts (highlighted by \textcolor{red}{$\times$}), wrong perspective relation as shown in the last two frames. Our approach, whether optimizing over the initial seed (SoS), selecting frame-by-frame along the temporal axis (SoP), or applying beam search (BS), consistently produces geometrically coherent videos with no visible artifacts (highlighted by \textcolor{green}{\checkmark}).}
\label{fig:tts_comparison}
\end{figure}

Budget scaling results are shown in Fig.~\ref{fig:budget_eval}. PSNR, SSIM, LPIPS, and RPT are for 3D consistency metrics, and SC, BC, MS, AQ, and IQ are the perceptual quality dimensions of VBench. We observe two findings. (1) \textbf{All variants exhibit scaling behavior}: as the budget increases, all three methods improve consistently on both geometric and perceptual metrics, confirming that our geometry-based reward offers a meaningful search signal. (2) \textbf{Budget configuration shapes the type of improvement}: SoS allocates $K>1$ parallel paths for greater diversity but lacks temporal refinement, achieving the lowest RPT yet weaker results elsewhere; SoP maintains $S>1$ seeds per step for stable fine-grained gains, yielding the best VBench trend; Beam Search combines both ($K>1$, $S>1$) and attains the strongest 3D reconstruction. Qualitatively (Fig.~\ref{fig:tts_comparison}), the baseline shows geometric artifacts and incorrect perspective in later frames, while all three strategies produce geometrically coherent videos throughout. Note that SoP shares the baseline's initial seed and thus an identical first frame, whereas SoS and Beam Search optimize the seed before generation and explore a larger space from the first frame.

\subsection{Post-hoc Alignment of Bidirectional Video Generation}
\label{sec:TTS_Post-hoc}

\subsubsection{Evaluation Setups.} We apply post-hoc alignment to a bidirectional DiT~\cite{wan2025wan} using our geometry-based reward, evaluating two complementary strategies: supervised fine-tuning (SFT) and Direct Preference Optimization (DPO). For both, we adopt Low-Rank Adaptation (LoRA)~\cite{hu2021lora} with rank $r=64$ and $\alpha=128$ applied to the q, k, v, and o projection modules of the DiT. For the auxiliary loss terms, we configure $\lambda=0.1$ and $\gamma=0.01$.

\subsubsection{Results of post-hoc alignment.} Quantitative results are shown in Tab.~\ref{tab:post-hoc}. SFT already yields notable improvements over the baseline across most geometric metrics, as fine-tuning on high-reward samples tilts the generation distribution toward geometrically preferred modes. DPO further improves upon SFT by explicitly contrasting winning and losing samples, achieving stronger geometric consistency across 3D reconstruction and multi-view metrics. Among DPO variants, our Reproj-Pts reward attains the best SSIM (0.7977), LPIPS (0.1789), and EPI (2.127), outperforming the Epipolar baseline on the metrics most reflective of dense geometric coherence. For overall video quality, both SFT and DPO consistently improve over the baseline on subject and background consistency, while our DPO (Reproj-Pts) achieves the best SC (97.05\%), BC (95.25\%), and IQ (76.63\%), demonstrating that geometric preference alignment enhances perceptual quality alongside geometric fidelity.

\FloatBarrier

\begin{table}[t!]
\centering
\caption{\textbf{Evaluation of Post-hoc Alignment}.}
\label{tab:post-hoc}
\setlength{\tabcolsep}{5.5pt}
\renewcommand{\arraystretch}{1.15}
\resizebox{\textwidth}{!}{%
\begin{tabular}{l | ccc ccc | cccccc}
\toprule
\multirow{2}{*}{\tightbold{Method}}
& \multicolumn{6}{c|}{\textit{Geometric Consistency Evaluation}} &
\multicolumn{6}{c}{\textit{Overall Video Quality Evaluation (VBench \%)}} \\
\cmidrule(lr){2-7} \cmidrule(lr){8-13}&
\tightbold{PSNR} $\uparrow$ & \tightbold{SSIM} $\uparrow$ & \tightbold{LPIPS} $\downarrow$ &
\tightbold{EPI} $\downarrow$ & \tightbold{RPX} $\downarrow$ & \tightbold{RPT} $\downarrow$ &
\tightbold{SC} $\uparrow$ & \tightbold{BC} $\uparrow$ & \tightbold{MS} $\uparrow$ &
\tightbold{DD} $\uparrow$ & \tightbold{AQ} $\uparrow$ & \tightbold{IQ} $\uparrow$ \\
\midrule
Baseline~\cite{wan2025wan} &
22.45 & 0.7548 & 0.2243 & 2.832 & \tightbold{0.998} & 1.783 &
95.98 & 94.43 & 97.61 & \tightbold{27.34} & \underline{57.74} & 76.30 \\
+ SFT (Reproj-Pts) &
23.52 & 0.7927 & 0.1842 & 2.337 & \underline{1.003} & 2.257 &
96.97 & 95.15 & \tightbold{97.96} & 15.62 & 57.73 & \underline{76.58} \\
+ SFT + DPO (Epipolar) &
\tightbold{23.57} & \underline{0.7973} & \underline{0.1818} & \underline{2.187} & 1.018 & \tightbold{1.385} &
\underline{96.98} & \underline{95.16} & \underline{97.88} & 25.39 & \tightbold{57.76} & 76.52 \\
\rowcolor{lightgray}+ SFT + DPO (Reproj-Pts) &
\underline{23.54} & \tightbold{0.7977} & \tightbold{0.1789} & \tightbold{2.127} & 1.022 & \underline{1.424} &
\tightbold{97.05} & \tightbold{95.25} & 97.77 & \underline{25.78} & 57.71 & \tightbold{76.63} \\
\bottomrule
\end{tabular}%
}
\end{table}

\subsection{Ablation Studies}

\subsubsection{Regularization in Post-hoc Alignment.}
The geometry reward inherently favors static frames, leading to dynamic degree (DD) collapse under DPO. The auxiliary loss in Eq.~\ref{eq:aux}, with static penalty $\lambda$ and smoothness weight $\gamma$, counteracts this collapse. Tab.~\ref{tab:eq8_ablation} reports deltas w.r.t.\ $\lambda=0,\gamma=0$, isolating the DD recovery attributable to Eq.~\ref{eq:aux} from the DPO objective itself. A moderate $\lambda$ restores DD with negligible quality cost, whereas an overly large $\gamma$ harms motion smoothness.

\begin{table}[t!]
\centering
\caption{\textbf{Ablation of Regularization Terms in Eq.~\ref{eq:aux}}.}
\label{tab:eq8_ablation}

\setlength{\tabcolsep}{5.5pt}
\renewcommand{\arraystretch}{1.15}

\resizebox{\textwidth}{!}{%
\begin{tabular}{c @{\hspace{14pt}} c}
\begin{tabular}{l cccccc}
\toprule
$\lambda$ & \tightbold{SC} $\uparrow$ & \tightbold{BC} $\uparrow$ & \tightbold{MS} $\uparrow$ & \tightbold{DD} $\uparrow$ & \tightbold{AQ} $\uparrow$ & \tightbold{IQ} $\uparrow$ \\
\midrule
$0.1$  & $+0.02$ & $-0.04$ & $+0.00$ & $+0.78$ & $-0.02$ & $+0.02$ \\
$0.5$  & $+0.05$ & $-0.08$ & $+0.02$ & $+0.78$ & $-0.06$ & $-0.01$ \\
$1.0$  & $-0.00$ & $-0.02$ & $+0.00$ & $+0.39$ & $-0.15$ & $-0.00$ \\
$10.0$ & $-0.02$ & $+0.17$ & $+0.08$ & $\tightbold{+1.56}$ & $-0.01$ & $-0.18$ \\
\bottomrule
\end{tabular}
&
\begin{tabular}{l cccccc}
\toprule
$\gamma$ & \tightbold{SC} $\uparrow$ & \tightbold{BC} $\uparrow$ & \tightbold{MS} $\uparrow$ & \tightbold{DD} $\uparrow$ & \tightbold{AQ} $\uparrow$ & \tightbold{IQ} $\uparrow$ \\
\midrule
$0.1$  & $-0.03$ & $+0.04$ & $-0.01$ & $-0.39$ & $+0.01$ & $+0.04$ \\
$0.5$  & $-0.04$ & $+0.05$ & $-0.02$ & $+0.00$ & $+0.04$ & $+0.05$ \\
$1.0$  & $+0.01$ & $+0.03$ & $-0.03$ & $+0.00$ & $-0.03$ & $+0.02$ \\
$10.0$ & $-0.05$ & $-0.10$ & $-0.12$ & $+0.00$ & $-0.12$ & $+0.02$ \\
\bottomrule
\end{tabular}
\end{tabular}}
\end{table}

\subsubsection{Geometry-Aware Sampling (GAS).}
Tab.~\ref{tab:gas_ablation} ablates the sampling strategy, patch size $p$, and ratio $\tau$, with deltas relative to our default (shallow attention, $p=4$, $\tau=20\%$). Attention-based selection clearly surpasses uniform and saliency sampling, and the default configuration of $p$ and $\tau$ best balances geometric coverage against perceptual quality.

\begin{table}[t!]
\centering
\caption{\textbf{Ablation of Geometry-Aware Sampling (GAS)}.}
\label{tab:gas_ablation}

\setlength{\tabcolsep}{5.5pt}
\renewcommand{\arraystretch}{1.15}

\resizebox{\textwidth}{!}{%
\begin{tabular}{l cccccc @{\hskip 14pt} l cccccc}
\toprule
\tightbold{Setting} & \tightbold{SC} $\uparrow$ & \tightbold{BC} $\uparrow$ & \tightbold{MS} $\uparrow$ & \tightbold{DD} $\uparrow$ & \tightbold{AQ} $\uparrow$ & \tightbold{IQ} $\uparrow$ &
\tightbold{Setting} & \tightbold{SC} $\uparrow$ & \tightbold{BC} $\uparrow$ & \tightbold{MS} $\uparrow$ & \tightbold{DD} $\uparrow$ & \tightbold{AQ} $\uparrow$ & \tightbold{IQ} $\uparrow$ \\
\midrule
Uniform     & $-0.06$ & $-0.00$ & $+0.08$ & $-4.30$ & $-0.11$ & $+0.02$ & Saliency     & $-0.17$ & $+0.04$ & $+0.08$ & $-4.69$ & $+0.02$ & $-0.16$ \\
Deep Attn   & $-0.12$ & $+0.17$ & $+0.05$ & $-2.34$ & $-0.26$ & $-0.19$ & $p{=}2$      & $-0.27$ & $-0.37$ & $-0.06$ & $+2.34$ & $+0.30$ & $+0.11$ \\
$p{=}8$     & $-0.02$ & $-0.02$ & $+0.01$ & $-1.17$ & $-0.02$ & $+0.02$ & $p{=}16$     & $-0.07$ & $+0.03$ & $-0.03$ & $-0.39$ & $+0.15$ & $-0.04$ \\
$\tau{=}5\%$ & $-0.18$ & $-0.32$ & $-0.06$ & $+1.17$ & $+0.25$ & $+0.11$ & $\tau{=}10\%$ & $-0.22$ & $-0.26$ & $-0.04$ & $+1.56$ & $+0.19$ & $+0.08$ \\
$\tau{=}40\%$& $-0.08$ & $-0.03$ & $-0.01$ & $-0.78$ & $+0.03$ & $-0.07$ & $\tau{=}80\%$ & $-0.08$ & $-0.03$ & $-0.01$ & $-0.78$ & $+0.03$ & $-0.07$ \\
\bottomrule
\end{tabular}}
\end{table}

\section{Conclusions}
In this paper, we introduced VIGOR, a Video Geometry-Oriented Reward framework that addresses the lack of explicit geometric supervision in current video diffusion models. Leveraging pretrained geometric foundation models, we proposed a physically grounded reward based on pointwise cross-frame reprojection error, coupled with a geometry-aware sampling strategy for robustness in low-texture and non-semantic regions. Extensive experiments show that VIGOR mitigates common temporal artifacts such as object deformation, spatial drift, and depth violations. We further validated its versatility across post-training alignment and inference-time optimization for both bidirectional and causal architectures, providing a practical and scalable solution for geometrically consistent video generation.

\section*{Acknowledgements}
This work was supported by the National Natural Science Foundation of China (Grant No. 62302240), the Beijing Major Science and Technology Project (Grant No. Z251100007125021), and by Hi! PARIS through the Hi! PARIS Chair 2024
held at École Polytechnique (LIX). Additional computational resources were provided by the Supercomputing Center of Nankai University and the IDRIS High-Performance Computing facilities (under allocation 2026-AD011014300R3, courtesy of GENCI).

\clearpage

%
%
\bibliographystyle{splncs04}
\bibliography{reference}

\clearpage
\appendix

\end{document}